 \documentclass[pmlr,twocolumn,10pt]{jmlr} 

\usepackage{times}
\usepackage{soul}
\usepackage{url}
\usepackage{hyperref}
\usepackage{caption}
\usepackage{subcaption}
\usepackage{graphicx}
\usepackage{amsmath}
\usepackage{booktabs}
\usepackage{algorithm}
\usepackage{algpseudocode}

\usepackage{multirow}
\usepackage{bbm}
\usepackage{mathrsfs}
\usepackage{amssymb,amsmath}

\usepackage{color}
\usepackage{xcolor}
\usepackage[switch]{lineno}
\usepackage{float}

\usepackage{booktabs}
\usepackage{siunitx}

\usepackage[switch]{lineno}

\theorembodyfont{\upshape}
\theoremheaderfont{\scshape}
\theorempostheader{:}
\theoremsep{\newline}

\jmlrvolume{LEAVE UNSET}
\jmlryear{2024}
\jmlrsubmitted{LEAVE UNSET}
\jmlrpublished{LEAVE UNSET}
\jmlrworkshop{Machine Learning for Health (ML4H) 2024} 

\title{MIK: \underline{M}odified \underline{I}solation \underline{K}ernel for Biological Sequence Visualization, Classification, and Clustering}

 \author{%
    \Name{Sarwan Ali} \Email{sali85@student.gsu.edu} \\
  \addr Georgia State University Atlanta GA, USA 
  \AND
  \Name{Prakash Chourasia} \Email{pchourasia1@student.gsu.edu}\\
  \addr Georgia State University Atlanta GA, USA 
  \AND
\Name{Haris Mansoor} \Email{16060061@lums.edu.pk}\\
\addr Lahore University of Management Science Lahore Pakistan
\AND
\Name{Bipin koirala} \Email{bkoirala3@gatech.edu}\\
\addr Georgia Institute of Technology Atlanta GA, USA
\AND
\Name{Murray Patterson} \Email{mpatterson30@gsu.edu}\\
  \addr Georgia State University Atlanta GA, USA 
 }
\begin{document}

\maketitle

\begin{abstract}
The t-Distributed Stochastic Neighbor Embedding (t-SNE) has emerged as a popular dimensionality reduction technique for visualizing high-dimensional data. It computes pairwise similarities between data points by default using an RBF kernel and random initialization (in low-dimensional space), which successfully captures the overall structure but may struggle to preserve the local structure efficiently. This research proposes a novel approach called the Modified Isolation Kernel (MIK) as an alternative to the Gaussian kernel, which is built upon the concept of the Isolation Kernel. MIK uses adaptive density estimation to capture local structures more accurately and integrates robustness measures. 
It also assigns higher similarity values to nearby points and lower values to distant points.
Comparative research using the normal Gaussian kernel, the isolation kernel, and several initialization techniques, including random, PCA, and random walk initializations, are used to assess the proposed approach (MIK). Additionally, we compare the computational efficiency of all $3$ kernels with $3$ different initialization methods. Our experimental results demonstrate several advantages of the proposed kernel (MIK) and initialization method selection. It exhibits improved preservation of the local and global structure and enables better visualization of clusters and subclusters in the embedded space. 
These findings contribute to advancing dimensionality reduction techniques and provide researchers and practitioners with an effective tool for data exploration, visualization, and analysis in various domains.
\end{abstract}




\section{Introduction}
Dimensionality reduction is used to deal with high-dimensional (HD) datasets, get rid of pointless or unnecessary features, and increase computing effectiveness, comprehending the neighborhood and geometrical structures of datasets,  which are the motivation among many. We look at common techniques, such as linear methods such as Principal Component Analysis (PCA)~\citep{wold1987principal1} and nonlinear methods (such as t-SNE), and analyze their benefits and drawbacks. Among many tasks, it is very popular to visualize HD data, especially in the natural sciences~\citep{linderman2019clustering}. The data can be visualized to help with cluster structure understanding or to help with distributional attributes intuition.
The ability to display and understand HD data in a lower-dimensional (LD) space has made dimensionality reduction increasingly popular in recent years~\citep{wang2021understanding}. A popular technique for visualizing complicated datasets is t-SNE (t-Distributed Stochastic Neighbor Embedding)~\citep{van2008visualizing,tang2016visualizing}. The process of t-SNE involves (random) initialization of the 2-dimensional vector along with a similarity matrix (kernel matrix computed using Gaussian kernel by default)~\citep{zhu2021improving}. Although there are other possible initializations used by researchers including PCA, and Random Walk~\citep{berti2015veracity}. The 2-dimensional matrix is iteratively modified using many iterations unless the quality of t-SNE converges. The kernel function selection has significant effects on how well t-SNE performs. The Gaussian kernel has some drawbacks even if it frequently performs well. Since the data is prone to being overly smoothed, it is challenging to keep both local and global structures preserved. This restriction may result in suboptimal visualizations and hinder the understanding of the findings.

The nonlinear dimensionality reduction method known as t-SNE (t-Distributed Stochastic Neighbor Embedding) was designed for the visualization of HD data~\citep{wattenberg2016use}. The foundation and operating principle of t-SNE are to maintain neighborhood relationships in the HD space and map them to an LD space while keeping their proximity~\citep{lambert2022globally}. In determining the degree of similarity between data points and creating the t-SNE embedding, probability distributions are used~\citep{arora2018analysis}. The computation of pairwise similarities, the formulation of the joint probability distribution, and the optimization process by minimizing the Kullback-Leibler divergence are all components of the algorithm~\citep{arora2018analysis}.
In t-SNE, pairwise similarities are typically computed using the Gaussian kernel. However, it has certain limitations like the Gaussian kernel tends to oversmooth the data, hiding local features and blurring together boundaries between clusters~\citep{zhang2021out}. Although it partially succeeded, the distinct HD clusters tended to overlap in the embedding, which is known as the ''crowding problem". The loss of significant information occurs when the intricate relationships present in the data are not captured by the kernel~\citep{kobak2020heavy}. 
We consider the idea of the isolation kernel as a substitute strategy to solve the drawbacks of the Gaussian kernel. Instead of using a fixed kernel width, the isolation kernel compares data points based on how isolated they are from one another~\citep{zhu2021improving,zhang2021out}. The advantage of the isolation kernel is its capacity to capture local structures, which are frequently significant in many real-world datasets.

Although the isolation kernel has a lot of potential advantages, it also has certain drawbacks.
The isolation kernel might provide erroneous connections between points and is vulnerable to noisy data. By introducing superior data dependencies and a really single global parameter-only approach, the Isolation kernel enhances visibility and eliminates the need to modify bandwidths~\citep{zhu2019improving}. There is unquestionably a trade-off between retaining global linkages and capturing local structure, and the isolation kernel may tend to emphasize local structures at the expense of the overall data structure~\citep{zhu2021improving,zhu2019improving}. There is an evident need for adjustment to improve the performance of t-SNE given the drawbacks of both the Gaussian and isolation kernels. 

In order to balance the preservation of local and global structures and prevent significant global patterns from being obscured by local changes, we propose to modify the version of the isolation kernel. We demonstrate in this paper that the modified isolation kernel is able to produce visual representations that are more trustworthy and insightful, allowing researchers and practitioners to acquire deeper insights into complex datasets.  
Additionally, we evaluate various initialization methods including Random, PCA, and Random Walk for all kernels discussed. This advancement will benefit various domains such as bioinformatics, where visualizing complex data is crucial for gaining insights and making informed decisions. The preservation of local structure and outlier handling in t-SNE visualizations is enhanced by the MIK's incorporation of resilience and adaptive density estimation techniques. The MIK allocates higher similarity values to nearby points and lower values to distant locations to preserve the local structure. This allows for more accurate representations of the data distribution in the t-SNE visualization.
Our contributions to this paper are as follows:
\begin{enumerate}
    \item The research proposes the Modified Isolation Kernel (MIK), as an alternative to the Gaussian and Isolation kernel. Which is intended to address the existing shortcomings in preserving local and global structures, and handling noisy data and outliers.

    \item We conduct a comparative study between the MIK, Isolation, and Gaussian kernel on various real-world biological datasets. To the best of our knowledge, no previous evaluation of such data (particularly for the isolation kernel) has been performed.

    \item Performance of the MIK is evaluated using a variety of initialization techniques, such as random, PCA-based, and random walk-based initialization. To the best of our knowledge, the random walk-based initialization for such biological data is not been explored in the literature (particularly for the isolation kernel).

    \item MIK contributes to the advancement of dimensionality reduction in the context of t-SNE. Enhancing the preservation of local structure and robustness to outliers boosts data exploration and analysis capabilities.

    \item MIK provides researchers and practitioners with a powerful tool for data exploration, visualization, and analysis across various fields. The proposed approach evaluates different kernels and initialization methods for t-SNE.


\end{enumerate}

\section{Related Work}
\label{sec_relatedWork}
In this section, we review the literature and studies conducted. We categorize the related work into three main areas: (1) Dimensionality Reduction Techniques, (2) t-SNE Variants and Extensions, and (3) Kernel-based Dimensionality Reduction Methods.

Dimensionality reduction is a crucial step in obtaining the most important features from complex data~\citep{gisbrecht2015parametric}. 
Transcriptomic datasets and other sizable high-dimensionality datasets are amenable to the application using the t-SNE.  Several alternative methods for dimensionality reduction are well-liked by academics, such as Principal Component Analysis (PCA)~\citep{wold1987principal1} and Multidimensional Scaling (MDS)~\citep{torgerson1952multidimensional}. These methods can preserve the global structure. In contrast, Isomap~\citep{tenenbaum2000global}, Laplacian Eigenmaps~\citep{belkin2001laplacian}, and more recent techniques that emphasize local structure. 
Other methods like Linear Discriminant Analysis (LDA)~\citep{fisher1936use} and UMAP~\citep{mcinnes2018umap}
have shown impressive visualization performance on a variety of real-world datasets. 
The trade-off between the preservation of local and global structure has always been a source of conflict for these methods because they can only handle one or the other, not both. In a recent research~\cite {linderman2019clustering}, the theoretical analysis of t-SNE was specifically taken into consideration. They demonstrated that when t-SNE is run with early exaggeration, points from the same cluster move toward one another which shrinks the embedding of any clusters leading to a bad visualization. 
Authors in~\citep{bibal2023dt} propose Barnes-Hut SNE, which builds vantage-point trees as a parsing approximation to the similarities between input objects and uses this approximation to the t-SNE gradient. Fast interpolation-based t-SNE (FIt-SNE)~\citep{linderman2019fast} uses polynomial interpolation and further accelerates using the fast Fourier transform (FFT). LargeVis~\citep{tang2016visualizing}, a method that builds an approximation of the K-nearest neighbor graph from the data before arranging the graph in the low-dimensional space. 
The UMAP~\citep{mcinnes2018umap}, TriMap~\citep{wang2021understanding} and Dt-SNE~\citep{bibal2023dt} methods are also popular but with conflicting measures such as not preserving global structure efficiently.
The kernel-based methods that map the data into a high-dimensional feature space, can also be used as part of visualization pipeline~\citep{gisbrecht2015parametric,kobak2020heavy}. There are many popular techniques such as Kernel PCA~\citep{hoffmann2007kernel}, Kernel t-SNE~\citep{gisbrecht2015parametric}, bi-kernel t-SNE~\citep{zhang2021out}, which leverage kernel functions to capture non-linear relationships in the data.



\section{Proposed Approach}
\label{sec_proposedApproach}
In this section, we first define the notation and present our proposed approach for enhancing the t-SNE algorithm using a modified isolation kernel. We discuss the limitations of the Gaussian kernel and the isolation kernel separately and then introduce our modifications to address these limitations.

Suppose a dataset $X=\{ x_1,x_2,...,x_n\}$ in $\mathbb{R}^d$, and a dataset $Y=\{ y_1,y_2,...,y_n\}$, the objective of t-SNE is to map $X \in \mathbb{R}^d$ to $Y \in \mathbb{R}^{d'}$, such that $d' < d$ and the similarity between points are preserved as much as possible. As t-SNE is mostly used for data visualization, the value of low dimensional space is normally selected as $d'=2$ or $3$. The similarity between a pair of points $x_i, x_j$ in the higher dimensional space is represented by a probability $P_{ij}$, similarly for low dimensional space points $y_i,y_j$ is represented by $Q_{ij}$. These similarities are computed based on a distance metric in respective spaces. The goal of t-SNE is to map points from $X$ to $Y$ such that the probability distribution between $P_{ij}$ and $Q_{ij}$ are as close as possible.

\subsection{Modified Isolation Kernel (MIK)}
Modified Isolation Kernel (MIK), denoted as:
\begin{equation}\label{eq_mik}
    \begin{aligned}
    k_{iso-mod}(x_i, x_j) = \frac{1}{\sqrt{p_i p_j}} \exp \bigg(-\frac{|x_i - x_j|^2}{2 \sigma_i \sigma_j}\bigg)\cdot \\
    \bigg(1 - \frac{n_i}{\sum_{k \neq i}n_k} \bigg) \cdot \bigg(1 - \frac{n_j}{\sum_{k \neq j}n_k}\bigg)
  \end{aligned}
\end{equation}

where $p_i$ and $p_j$ are the adaptive density estimate of data points indexed at $i$ and $j$ respectively. $x_i$ and $x_j$ represent data points in the input space and $\sigma_i, \sigma_j$ correspond to the standard deviation computed based on the adaptive neighborhood size of the points. $n_i \text{ and } n_j$ are the local densities of the input points.

\begin{remark}
Let, $x_1, \cdots, x_n \in \chi \subset \mathbb{R}^d$ be a finite-collection of data-points in the input space. A kernel $k(\cdot, \cdot): \chi ^2 \rightarrow \mathbb{R}$ is said to be positive semi-definite, if the Gram matrix $K \in \mathbb{R}^{n \times n} - $ defined by $K_{ij} = k(x_i, x_j)$ is positive semi-definite.\\
\end{remark}

$k(\cdot, \cdot): \chi^2 \rightarrow \mathbb{R}$ is a valid kernel if it is;
\begin{itemize}
    \item Symmetric
    \item Positive semi-definite
\end{itemize}

\paragraph{MIK is Symmetric: }
For a kernel to be valid, it must be symmetric. Symmetry implies $$k_{iso-mod}(x_i, x_j) = k_{iso-mod}(x_j, x_i); \quad \forall \  (i, j)$$

The first term, $\frac{1}{\sqrt{p_i p_j}}$, is symmetric because the product is commutative. The exponential term $\exp \big(-\frac{|x_i - x_j|^2}{2\sigma_i \sigma_j}\big)$ is symmetric because $|x_i - x_j| = |x_j - x_i|$ and the product of the standard deviations $\sigma_i, \sigma_j$ is also commutative. Furthermore, the multiplicative terms $\big(1 - \frac{n_i}{\sum_{k \neq i}n_k}\big)$ and $\big(1 - \frac{n_j}{\sum_{k \neq j}n_k}\big)$ are independent of the other index and will simply swap when $i$ and $j$ are exchanged.\\

Thus, each part of the expression in $K_{iso-mod}(x_i, x_j)$ is symmetric with respect to $i$ and $j$, proving that the proposed kernel is symmetric.

\paragraph{Positive Semi Definite (P.S.D)}
A kernel can be thought of as an inner product in some feature space, i.e. there exists a mapping $\phi: \mathcal{X} \rightarrow \mathcal{H}$, where $\mathcal{H}$ is a Hilbert space, such that $k(x_i, x_j) = \langle \phi(x_i), \phi(x_j)\rangle$. This ensures that the Gram matrix $K$ is positive semi-definite.\\

The Gaussian component, $\exp \big(-\frac{|x_i - x_j|^2}{2 \sigma_i \sigma_j}\big)$, in $(1)$ is a well-known \textit{Mercer Kernel}~\citep{mercer1909xvi} (RBF kernel) and is positive semi-definite by itself and it can be expressed as an inner product in an infinite-dimensional feature space i.e. $\langle \phi_{gauss}(x_i), \phi_{gauss}(x_j)\rangle$. The scaling factor $\frac{1}{\sqrt{p_i p_j}}$ is a positive quantity and multiplying a Mercer kernel by a positive weight still preserves the kernel's positive semi-definiteness. This scaling term does not invalidate the kernel but introduces a density-based scaling in the feature space. We can interpret this as scaling the feature space representation:
$$\phi_{scaled}(x_i) = \frac{1}{\sqrt{p_i}} \phi_{gauss}(x_i)$$

The terms $\big(1 - \frac{n_i}{\sum_{k \neq i}n_k}\big)$ and $\big(1 - \frac{n_j}{\sum_{k \neq j}n_k}\big)$ represent local density corrections based on DBSCAN density estimates $n_i$ and $n_j$. These terms are bounded between 0 and 1, and they modify the similarity based on how densely packed points are in the neighborhood of $x_i$ and $x_j$, and can also be interpreted as a scaling of the kernel in the feature space. The final feature space represented by the \textit{Modified Isolation Kernel} (MIK) is $$\phi_f(x_i) = \bigg(1 - \frac{n_i}{\sum_{k \neq i}n_k}\bigg) \phi_{scaled}(x_i)$$\\

The kernel now can be expressed as; $ k_{iso-mod}(x_i, x_j) = \langle \phi_f{x_i}, \phi_f{x_j}\rangle$. Thus, MIK is a valid kernel function.\\

Furthermore, the quantity $\frac{1}{\sqrt{p_i p_j}}$ has a finite upper bound, say $U_{density}$, which depends on the dataset's density distribution and the entire kernel function is bounded as $$k_{iso-mod}(x_i, x_j) \leq U_{density}$$

Since, $k_{iso-mod}(x_i, x_j)$ is continuous symmetric positive semi-definite kernel which is bounded, Mercer's Theorem~\citep{mercer1909xvi} reveals that the kernel function can be expressed as; $$k_{iso-mod}(x_i, x_j) = \sum_{k=1}^\infty \lambda_i \psi_k(x_i) \psi_k(x_j)$$ where $\{\lambda\}_{k=1}^\infty$ are non-negative sequence of eigen-values and $\{\psi (\cdot)\}_{k=1}^\infty$ are the corresponding orthonormal bases in $\mathcal{L}^2(\mathcal{X})$. The convergence of this series is absolute and uniform.\\

Together, these properties ensure that the kernel function can be expressed in a stable, well-behaved manner and that it will lead to positive semi-definite kernel matrices for any dataset, which is crucial for the practical application of kernel-based methods.

\subsection{t-SNE with Modified Isolation Kernel (MIK)}
The modified isolation kernel is designed to improve the robustness, balance, and interpretability of t-SNE visualizations. It achieves this by incorporating adaptive neighborhood weights and local density information into the kernel computation. It uses the same Algorithm~\ref{algo_tsne}. The only difference is while computing the affinity matrix here we use a Modified Isolation kernel (MIK). 

\textbf{Adaptive Neighborhood Weights: }
The modified isolation kernel incorporates a term $\frac{1}{\sqrt{p_i p_j}}$, where $p_i$ and $p_j$ are the adaptive density estimates of data points $i$ and $j$, respectively. This term adjusts the similarity computation based on the local densities of the data points, allowing for a more effective capture of the local structure. In contrast, the simple isolation kernel does not include a density-based formulation.

The terms $p_i$ and $p_j$ are computed using a weight term represented by $n_k$,
The weights ($n_k$) are computed using a density-based clustering algorithm called DBSCAN (Density-Based Spatial Clustering of Applications with Noise). The weights represent the density or importance of each data point in the computation of the similarity matrix. We can assign higher weights to densely populated regions and lower weights to sparser regions. This adaptive weighting scheme helps in capturing the local structures effectively. The step-by-step explanation for weights computation:
\begin{enumerate}
    \item \textbf{Create a DBSCAN object:} Instantiate a DBSCAN object with the desired parameters, such as the neighborhood radius (epsilon) and the minimum number of points required to form a dense region (min. samples).
    \item \textbf{Fit DBSCAN on data:} Apply the DBSCAN algorithm to the high-dimensional data X by calling the fit method of the DBSCAN object. This step identifies dense regions and assigns labels to each data point.
    \item \textbf{Get the labels:} Access the labels attribute of the DBSCAN object. This attribute contains the cluster labels assigned to each data point by the DBSCAN algorithm. The labels indicate whether a data point is a core, a border, or a noise point.
    \item \textbf{Assign weights:} Assign weights based on the DBSCAN labels. The weights can be assigned as follows:
    \begin{enumerate}
    \item For core points (DBSCAN label $>$ -1): Assign a weight of 1. These points are considered important for preserving the local structure.
    \item For border points (DBSCAN label = -1): Assign a weight of 0.5 (or any other appropriate value). These points are located near core points but do not have enough neighboring points to be classified as core points.
    \item For noise points (DBSCAN label = -2): Assign a 0 weight (or very small value). 
    These points are not within the neighborhood of any core point and are considered less important.
\end{enumerate}
\end{enumerate}

The weights variable (represented by $n_k$) will contain the computed weight corresponding to a data point in X based on the DBSCAN clustering. 
The values of $p_i$ and $p_j$ can be computed based on the weights ($n_k$) obtained from the DBSCAN algorithm. Here's how the values of $p_i$ and $p_j$ can be calculated:

\begin{enumerate}
    \item \textbf{Compute the sum of weights:} Calculate the sum of weights for all data points except the current data point $i$ (denoted as $\sum_{k\neq i} n_k$) and the sum of weights for all data points except the current data point $j$ (denoted as $\sum_{k\neq j} n_k$). This can be done by summing up the weights of all other data points in the respective groups.
    \item \textbf{Compute $p_i$:} For a given data point $i$, $p_i$ can be computed as the inverse of the square root of the product of its weight and the sum of weights of all other data points except $i$. Mathematically, $p_i = \frac{1}{\sqrt{n_i \left(\sum_{k\neq i} n_k\right)}}$.
    \item \textbf{Compute $p_j$:} Similarly, for a given data point $j$, $p_j$ can be computed as the inverse of the square root of the product of its weight and the sum of weights of all other data points except $j$. Mathematically, $p_j = \frac{1}{\sqrt{n_j \left(\sum_{k\neq j} n_k\right)}}$.
\end{enumerate}

These weights can then be used in the modified isolation kernel or any other relevant calculations. See Equation~\ref{eq_mik} for the definition of MIK. The detailed theoretical proofs for MIK are given in Section~\ref{sec_proof_appendix} (in the appendix). The details regarding tSNE with Gaussian and Isolation kernel are given in Section~\ref{methodology_appendix} in the appendix.

For t-SNE, we used different initialization methods, namely random-, PCA-, and random walk-based initialization. A detailed description of each initialization method is given below.

\subsection{Initialization Methods}\label{sec_init_appendix}
In the t-SNE algorithm, the initialization of the low-dimensional embedding plays a crucial role in determining the convergence and quality of visualization and dimensionality reduction. We explore three different methods for initializing the 2D t-SNE representation: Random, PCA-based, and random walk-based initialization.

\subsubsection{Random Initialization}
Low-dimensional embedding coordinates are randomly initialized. While the approach is straightforward, it may lead to suboptimal results as the initial positions are not informed by the underlying structure of high-dimensional data.

\subsubsection{PCA-based Initialization}

This method leverages the principal components of the high-dimensional data to initialize the 2D t-SNE representation, and principal component analysis (PCA) is performed to capture the major directions of variance in the data and provide a better starting point for the t-SNE optimization.

\subsubsection{Random Walk-based Initialization}

It incorporates the concept of random walks to determine the initial positions of the data points in the 2D space. This approach aims to preserve local neighborhood information and encourage the clustering of similar data points. It follows these steps: (1) Initialize the low-dimensional embedding, $Y$, with random values for each data point, (2) Perform a random walk for fixed steps (e.g., 1000 steps), (3) For each data point, select a random neighbor, and (4) Update the position of the current data point towards the selected neighbor by adjusting its coordinates based on the Euclidean distance and the number of steps taken.
By iteratively updating the positions of the data points based on random walks, this initialization method encourages nearby points to be closer to the 2D space, thereby preserving local structures. The algorithmic pseudocode for random walk-based initialization is given in Algorithm~\ref{algo_random_walk}.

\begin{algorithm2e}[h!]
\caption{Random Walk-based Initialization}
\label{algo_random_walk}
\scriptsize
 \LinesNumbered
\KwIn{Number of data points $n$, number of dimensions $no\_dims$}
\KwOut{Low-dimensional embedding $Y$}
Initialize $Y$ with random values for each data point;

\For{$i = 1$ \textbf{to} $n$}{

    \tcc{Below is a loop for the Number of random walk steps}
    
    \For{$j = 1$ \textbf{to} $1000$ }
    {  
        Randomly select a neighbor $neighbor$ from $n$ data points
        
        
        $Y[i] += \frac{Y[neighbor] - Y[i]}{\sqrt{j + 1}}$
    }
}
\textbf{return} $Y$
\end{algorithm2e}

\subsection{Embedding Methods}\label{sec_embed_appendix}

\paragraph{\textbf{Spike2Vec}~\citep{ali2021spike2vec}} Using the idea of $k$-mers (also known as n-grams), it produces fixed-length numerical representations. It creates mers, or substrings, of length $k$ (the window size), using the idea of a sliding window for sequence.

\paragraph{\textbf{Spaced $k$-mer}~\citep{singh2017gakco}} 
It computes a gapped version of $k$-mers by first computing a $g$-mers, where $g>k$. Then the $k$-mer is extracted from $g$-mer, which contains the first $k$ amino acids/nucleotides in lexicographically sorted order in both forward and reverse direction. Embedding is then generated based on the count of these gapped $k$-mers.

\paragraph{\textbf{PWM2Vec}~\citep{ali2022pwm2vec}} It follows the fundamental principles of $k$-mers spectrum, but instead of using fixed frequency values, it assigns a weight to each amino acid in the $k$-mers using position weight matrix (PWM).

\subsection{t-SNE Algorithm and Workflow}
\begin{figure}[h!]
    \centering
    \includegraphics[scale=0.24]{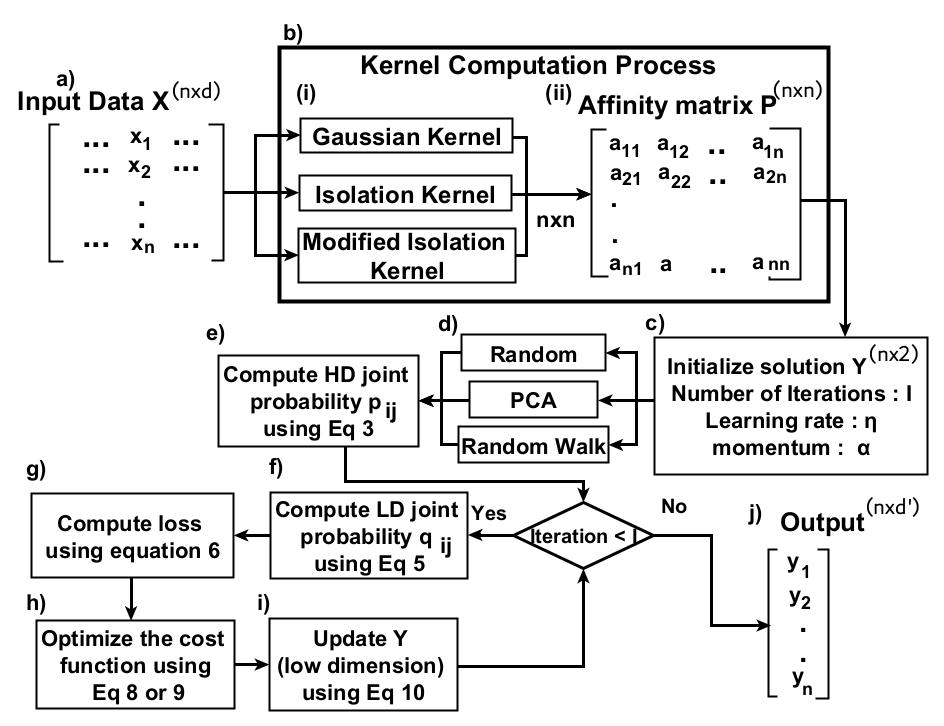}
    \caption{Workflow - We initially construct the kernel matrix based on the input data utilizing different kernel functions. To get the low-dimensional representation of the data, we next employ the t-SNE computing process. 
    }
    \label{process_flow_chart}
\end{figure}

Figure~\ref{process_flow_chart} shows the flow chart for t-SNE which is also illustrated in Algorithm~\ref{algo_tsne}. The first step is to calculate the pairwise affinities (similarities). This can be done using different kernel functions to get a similarity matrix including Gaussian, Isolation, and Modified Isolation Kernel as shown in Figure~\ref{process_flow_chart}-b)-(i). Now we initialize our solution $Y$ with different initialization methods including Random, PCA, or Random Walk as shown in Algorithm~\ref{algo_tsne} Line 2 also in Figure~\ref{process_flow_chart}-c) and d). Then we get the probability value from the distances for the points $X$ (where $X \in R^d$, In higher dimension) as shown in Algorithm~\ref{algo_tsne} Line 11 also in Figure~\ref{process_flow_chart}-e). The next step is to compute the joint probability in high dimension $P_{ij}$ and also in low dimension $Q_{ij}$ as shown in Algorithm~\ref{algo_tsne} Line 18 also in  Figure~\ref{process_flow_chart}-f). \textcolor{black}{We compute KL divergence to measure the variation in two distributions, this optimizes the computational cost as shown in Algorithm~\ref{algo_tsne} Line 21 also in Figure~\ref{process_flow_chart}-g)}. Computing derivative for this cost function and calculating gradient descent gives the optimal $Y$ after multiple iterations as shown in Algorithm~\ref{algo_tsne} Line 22 and 23 also in Figure~\ref{process_flow_chart}-h) and h). The updated $Y$ is the lower dimensional representation of high dimensional data points $X$ Figure~\ref{process_flow_chart}-j).

\begin{algorithm2e}[h!]
\caption{t-SNE Computation}
\label{algo_tsne}
\LinesNumbered
\scriptsize
\KwIn{$X$, affMat, d'}
\KwOut{$Y = [y_1, \ldots, y_{n}], y_i \in R^{d'}$}
Compute $\sigma_i$ for every $x_i \in X$	using perplexity
\SetKwFunction{FMain}{tSNE}
\FMain{$KM$, dim}{

    $Y =  matrix(n, dim)$ \Comment{initialize solution matrix}
        
    \tcc{Random, PCA or Random Walk Initialization }
    
    \tcc{Now initialize optimization parameters}
        
    $I$ = 1000 \Comment{Iterations}
        
    $\eta = 500$,  $\alpha = 0.5$ \Comment{Learning rate and momentum}
        
        
        $P = matrix(n \times n)$  \Comment{probability matrix in HD}
    
       \For{$i \leftarrow 1 \textup { to } n$}{ 
           \For{$j \leftarrow 1 \textup { to } n$}{
             $P_{ij}$ = \Call{computeProbability}{$affMat[i][j]$} \Comment{from Eq~\ref{eq_symmetric_conditional_HD} (in the appendix)}
          }
        }
        
        $Q = matrix(n \times n)$  \Comment{probability matrix in LD}

        \tcc{Now starting Iteration loop}
        
        \For{$k \leftarrow 1 \textup { to } I$}{  
        
         \For{$i \leftarrow 1 \textup { to } n$}{
           \For{$j \leftarrow 1 \textup { to } n$}
            {
            
            $Q_{ij}$ = \Call{computeProbability}{$Y_{i,j}$} \Comment{Eq.~\ref{eq_ld_jointprobability} (in the appendix)}
            }
        }

        $L = \sum_i \sum_j P_{ij} log \frac{P_{ij}}{Q_{ij}}$ \Comment{Loss computation}
        
        $\frac{\partial L}{\partial Y}$ = \Call{computeGradient}{$L$} \Comment{using Eq~\ref{eq_gradient} (in the appendix)}
        
        $Y$ = \Call{updateOutput}{$Y, \alpha, \eta, L$} \Comment{using Eq~\ref{eq_update_y} (in the appendix)}

        \If{($k$ == $250$)}{  
           
          $\alpha$ =  $0.8$   \Comment{momentum changed at iteration} 
        }
    }
    \Return $Y$          
}
\end{algorithm2e}

\section{Experimental Setup}
\label{sec_expereimentalSetup}
In this section, we describe the experimental setup used to evaluate the performance of the proposed MIK in t-SNE. We discuss the dataset, evaluation metrics, and parameter settings employed in our experiments. All the experiments are performed using Python on a Core i5 system with a $2.4 GHz$ processor having $32$ GB memory and Windows 10 OS. 


\paragraph{\textbf{tSNE Evaluation Metrics:}}
To assess the quality of the t-SNE visualizations, we employed two commonly used evaluation metrics, namely Neighborhood Agreement (NA) and Trustworthiness (TW). A detailed description of these methods along with the evaluation metrics for classification and clustering are given in Section~\ref{appendix_eval} in the appendix.

\paragraph{\textbf{Dataset Statistics:}}
We use $3$ biological datasets for experimentation Protein Subcellular~\citep{ProtLoc_website_url}, GISAID~\citep{gisaid_website_url}, and Nucleotide~\citep{human_dna_website_url}.
Their summary is shown in Table~\ref{tbl_data_statistics} (in appendix).

\begin{figure*}[h!]
  \centering
 \subfigure[Spike2Vec (Rand.)]{
 \includegraphics[scale=0.550]{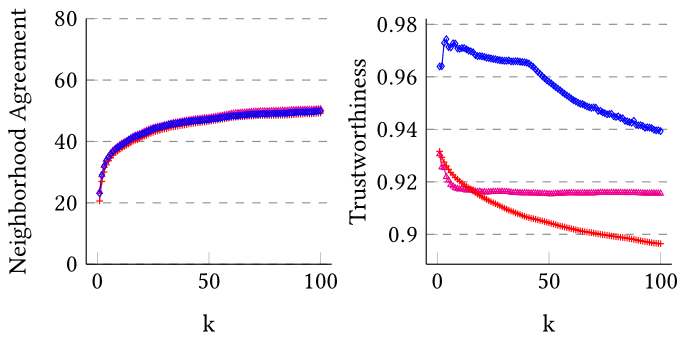}
    }%
  \subfigure[Spaced $k$-mers (Rand.)]{
  \includegraphics[scale=0.550]{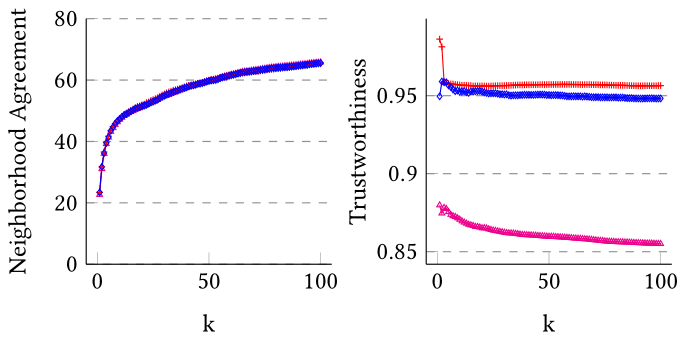}
  }%
\subfigure[PWM2Vec (Rand.)]{
  \includegraphics[scale=0.550]{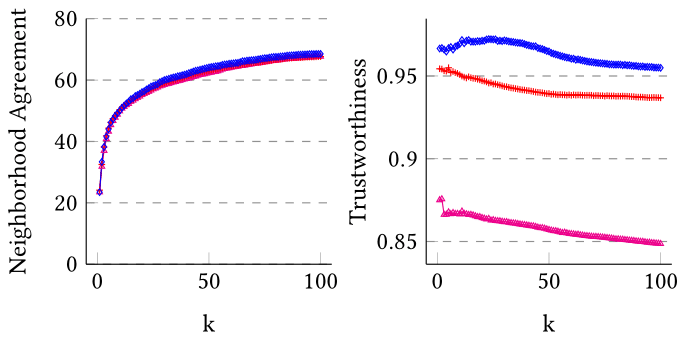}
  }%
  \\
 \subfigure[Spike2Vec (PCA)]{
  \includegraphics[scale=0.550]{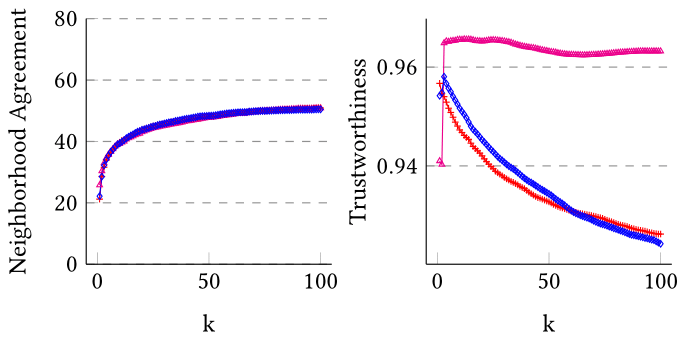}
  }%
 \subfigure[Spaced $k$-mers (PCA)]{
  \includegraphics[scale=0.550]{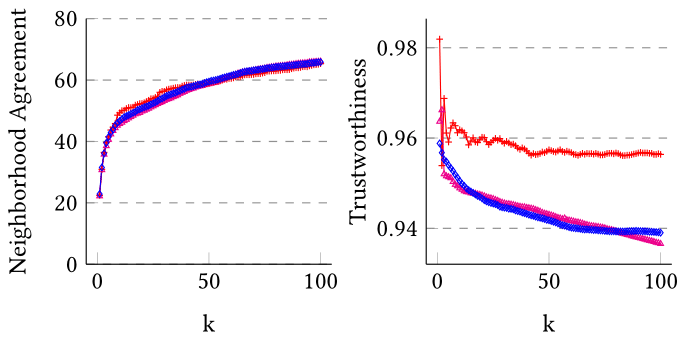}
  }%
 \subfigure[PWM2Vec (PCA)]{
  \includegraphics[scale=0.550]{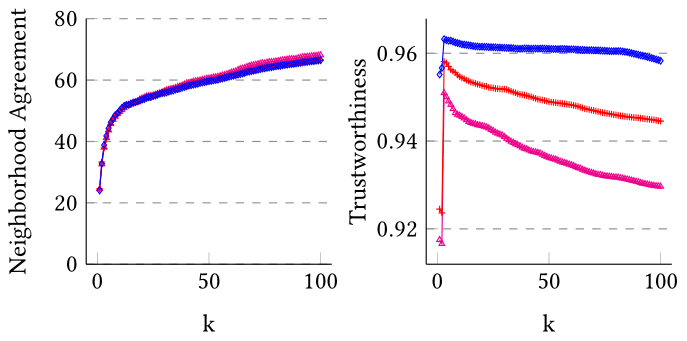}
  }%
  \\
  \subfigure[Spike2Vec (Walk)]{
  \includegraphics[scale=0.550]{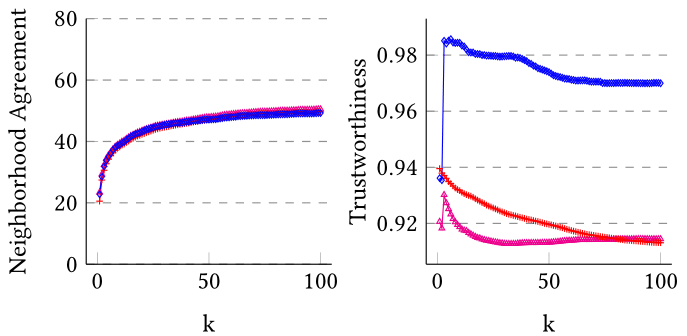}
  }%
 \subfigure[Spaced $k$-mers (Walk)]{
  \includegraphics[scale=0.550]{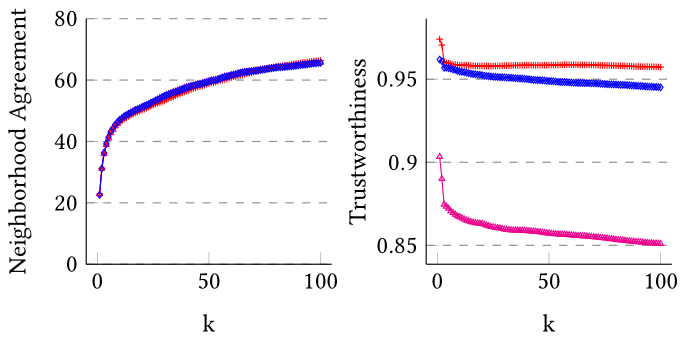}
  }%
  \subfigure[PWM2Vec (Walk)]{
  \includegraphics[scale=0.550]{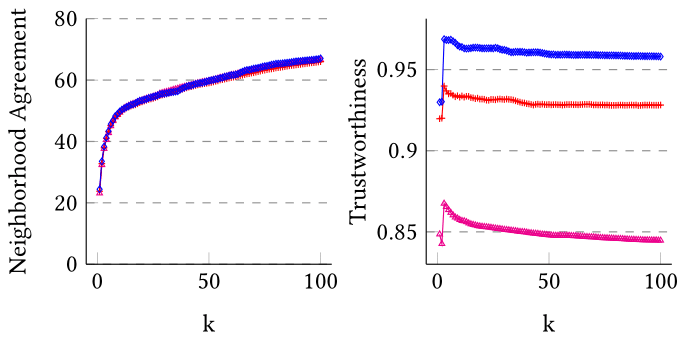}
  }%
  \\
  \includegraphics[scale=0.95]{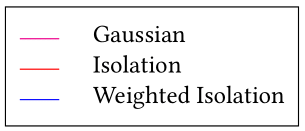}
  \caption{Neighborhood Agreement (left) and Trustworthiness (right) for tSNE-based 2D representation of \textbf{GISAID} Dataset using a different number of neighbors $k$  (in x-axis) for different Embedding  (see subcaptions). These results are computed for \textbf{random initialization} (top row), \textbf{PCA-based initialization} (middle row), and \textbf{Random Walk-based initialization} (bottom row) of tSNE. The pink lines show results for the Gaussian kernel, the red line is for Isolation, while the blue line is for our proposed weighted/modified isolation kernel. The figure is best seen in color.} 
  \label{fig_Auc_GISAID_spike2Vec_random_neighborhood}
\end{figure*}

\section{Results And Discussion}
\label{sec_resultsAndDiscussion}
In this section, we report the results for visualization, classification, and clustering.

\paragraph{\textbf{Visualization Results:}}
The neighborhood agreement and trustworthiness results for GISAID data with different initialization-based tSNEs are reported in Figure~\ref{fig_Auc_GISAID_spike2Vec_random_neighborhood} for Spike2Vec, Spaced $k$-mers, and PWM2Vec-based embeddings. 
Similarly, results for the trustworthiness of Nucleotide and Protein subcellular datasets are reported in Figures~\ref{fig_Auc_humanDNA_spike2Vec_random_neighborhood} and ~\ref{fig_Auc_Protein_Subcellular_spike2Vec_random_neighborhood} (in the appendix), respectively.
For the GISAID dataset, we can observe that in the case of neighborhood agreement, the proposed modified kernel either performs comparable or better than both Gaussian and Isolation kernels for all embedding and initialization methods. For neighborhood agreement, the proposed modified isolation kernel significantly outperforms for Spike2Vec (with random and random walk initialization) and PWM2Vec (with random, PCA, and Random walk initialization). In summary, the proposed method outperforms all other methods for PWM2Vec-based embedding while showing superior performance for Spike2Vec (in most cases).
For the Nucleotide dataset, we can observe that the proposed MIK is comparable with both Gaussian and isolation kernels in the case of neighborhood agreement while it performs better than the Gaussian kernel for trustworthiness. 
Similarly, in the case of the Protein Subcellular dataset, for the neighborhood agreement metric, we can observe that the performance of the proposed Modified Isolation kernel method is similar to that of the Gaussian kernel and Isolation kernel. However, for trustworthiness, the proposed method does not perform the best (although the performance difference is not very significant). This is because the separation in the protein subcellular data is not captured by the proposed method.
We also provide a discussion on the statistical significance of the t-SNE-based results in Section~\ref{sec_stats_tsne_appendix} in the appendix.

\paragraph{\textbf{Classification Results:}}
The classification results for the protein subcellular data are shown in Table~\ref{tbl_results_classification_protein_subcellular} (in the appendix). We can observe that the Spaced $k$-mers embedding with our Modified Isolation kernel (MLP and random forest classifiers) outperforms the Gaussian and Isolation kernel and all other embedding methods for all evaluation metrics other than training runtime (for which our proposed MIK with PWM2Vec performs the best using naive bayes classifier). Similar behavior can also be observed in the classification results for the GISAID dataset and Nucleotide dataset in Table ~\ref{tbl_results_classification_GISAID} (in the appendix) and Table~\ref{tbl_results_classification_Human_DNA} (in the appendix), respectively.
The behavior from classification results shows that the modified isolation kernel can compute the pairwise similarities between sequences in a better way compared to the Gaussian and Isolation kernels (in most cases). This is because MIK contains neighborhood weights embedded into the similarity values, hence performing better in this regard.
The discussion related to the statistical significance of classification results is shown in Section~\ref{sec_stats_classi_appendix} in the appendix.

\paragraph{\textbf{Clustering Results:}}
For clustering, we took the optimal number of clusters similar to the number of unique labels in the datasets (see $3rd$ column of Table~\ref{tbl_data_statistics} in the appendix), making it a data-driven approach. This was also confirmed by the elbow method~\citep{satopaa2011finding}.
The results for the Protein Subcellular dataset are reported in Table~\ref{tbl_results_clustering_protein_subcellular} (in the appendix). We can observe that for the Silhouette coefficient (SC), the isolation kernel (IK) with spaced $k$-mer using Agglomerative clustering (AC) outperforms all methods. However, the SC with MIK, spaced $k$-mers, and AC is comparable to IK. The MIK with Spike2Vec and K-means outperforms all methods for the Calinski score. Similarly, the IK performs better than others for Davies's score.
Since we used $1000$ iterations for clustering, the results are statistically significant and stable because of the robustness of random initialization and convergence assessment. Similar behavior can be observed for the clustering results for GISAID and Nucleotide datasets in Table~\ref{tbl_results_clustering_GISAID} and ~\ref{tbl_results_clustering_humna_dna} (in the appendix).

\paragraph{\textbf{Discussion:}}
Visualization, classification, and clustering results show that the proposed MIK can generalize better than Gaussian and Isolation kernel for all $3$ tasks, hence showing that by incorporating the neighborhood weights information within the pairwise similarities, we can preserve the similarities between data points efficiently. 
We also provide recommended initialization that is better for different datasets based on all the results reported above. Our suggestions are reported in Table~\ref{tbl_recommended_result_summary} (in the appendix), which states that in general, random walk-based initialization provides better results while simple random initialization, which is used as a default method for tSNE, provides the worst results. 
Although the proposed method does not significantly outperform the existing methods for all datasets and initialization methods, we believe that this type of analysis could be used as a benchmark for further analysis of data visualization methods (in the future).


\section{Conclusion}
\label{sec_conclusion}
We propose a method for the t-SNE computation, the MIK, as an alternative to the Gaussian kernel (GK). By adding adaptive density estimation to improve the preservation of local and global structures, the MIK solves the drawbacks of the GK and IK.
Experimental results on diverse datasets show that the MIK is superior to the others. 
Future research will focus on evaluating the MIK's performance with other datasets along with improving computational efficiency. 
Another exciting future extension is to compare results with other visualization methods like UMAP. 


\bibliography{references}

\clearpage


\appendix

\section{Methodology}
\label{methodology_appendix}
\subsection{t-SNE with Gaussian Kernel}



In t-SNE with the Gaussian Kernel~\citep{hinton2002stochastic,xue2020classification}, the similarity between $x_i$ and $x_j$ in high dimensional space is measured using a Gaussian kernel as given below: 
\begin{equation}\label{eq:1}
K(x_i,x_j)=exp(\frac{-|| x_i- x_j||^2}{2\sigma_{i}^{2}})
\end{equation}

The conditional probability ($P_{j \vert i}$) that $x_i$ would pick $x_j$ as its neighbour is given below: 
\begin{equation}\label{eq:2}
P_{j \vert i}=\frac{K(x_i,x_j)}{\sum_{k \neq i}^{} K(x_i,x_k)}
\end{equation}

The  symmetric conditional probability (similarity) between $x_i$ and $x_j$ in high dimensional space is represented by $P_{ij}$ is given as:
\begin{equation}
\label{eq_symmetric_conditional_HD}
P_{i j}= \begin{cases}0 & j=i \\ 
\frac{P_{i \vert j}+P_{j \vert i}}{2n} & j \neq i\end{cases}
\end{equation}

The similarity between $y_i$ and $y_j$ in low dimensional space is defined as:
\begin{equation}\label{eq:4}
K'(y_i,y_j)= \left(1+\left\|y_{i}-y_{j}\right\|^{2}\right)^{-1}
\end{equation}

The probability (similarity) between $y_i$ and $y_j$ in low dimensional space is represented by $Q_{ij}$ is given as:

\begin{equation}
\label{eq_ld_jointprobability}
Q_{i j}= \begin{cases}0 & j=i \\ \frac{\left(1+\left\|{y}_{i}-{y}_{j}\right\|^{2}\right)^{-1}}{\sum_{k \neq l}\left(1+\| {y}_{k}-{y}_{l}||^{2}\right)^{-1}} & j \neq i\end{cases}
\end{equation}

The distance-based similarity measure $K'$ is a heavy-tailed distribution i.e. it approaches an inverse square law for large pairwise distances. This means that far-apart points have $Q_{ij}$ invariant to changes in the scale of low dimensional space.

The t-SNE algorithm is designed to find the best value of $\sigma_i$ such that the perplexity of the conditional distribution equals a fixed value defined by the user. To accomplish this t-SNE performs a binary search. In this way, the bandwidth is adapted to the density of the data such that small values of $\sigma_i$ are used in dense data regions and large used in sparse regions respectively. The perplexity is a smooth measure of the effective number of neighbors like $k$ in the nearest neighbors. Each point in $y_i \in Y$ is computed by minimizing a cost function on the Kullback-Leibler (KL) divergence of the joint probability distribution in the low dimensional space ($Q$) from the joint distribution in high dimensional space $P$ as given in Equation~\eqref{eq_KL_div}.

\textbf{KL Divergence Loss:} The KL divergence (Kullback-Leibler divergence) is a measure of dissimilarity between two probability distributions. In t-SNE, the KL divergence is used as the objective function to minimize during the optimization process. The original formula for the KL divergence between two distributions P (high dimensional) and Q (low dimensional) is as follows:
\begin{equation}\label{eq_KL_div}
    KL (P \vert \vert Q) = \sum_i \sum_j P_{ij} log \frac{P_{ij}}{Q_{ij}}
\end{equation}

The goal is to find low-dimensional embeddings that minimize the KL divergence between P and Q.

\textbf{Compute Gradient: } The gradient $\frac{\partial L}{\partial Y}$ in the t-SNE algorithm is computed to optimize the embedding space. It represents the direction and magnitude of the change that needs to be made to the embedding coordinates to minimize the cost function $C$ and can be denoted by : 

\begin{equation}\label{eq_gradient}
\frac{\partial L}{\partial y_i} = 4 \sum_{i=1}^{n} \left(\sum_{j=1}^{n} P_{ij} - Q_{ij}\right) (y_i - y_j)
\end{equation}


where $N$ is the number of data points, $P_{ij}$ is the probability between points $i$ and $j$ in high dimensional space, $Q_{ij}$ is the probability between points $i$ and $j$ in the low dimensional space. $y_i$ and $y_j$ are the low-dimensional coordinates of points $i$ and $j$. 

The gradient $\frac{\partial L}{\partial Y}$ is computed using Equation~\eqref{eq_gradient}, which represents the gradient of the Kullback-Leibler divergence between $P$ and $Q$. Finally, we update the low-dimensional embedding $Y$ using gradient descent with momentum. The algorithm iterates for the specified number of iterations and returns the final low-dimensional embedding $Y$. The t-SNE uses the gradient descent method to solve the optimization objective problem by using Equation~\ref{eq_gradient}. The output is updated using Equation~\ref{eq_update_y}.

\begin{equation} \label{eq_update_y}
y^{(t)}=y^{(t-1)}+\eta \frac{\partial L}{\partial Y} +\alpha(t)\left(y^{(t-1)}-y^{(t-2)}\right)
\end{equation}


\subsection{t-SNE with Isolation Kernel}

The Isolation kernel~\citep{zhu2021improving} is an alternative to the Gaussian kernel for similarity computation in t-SNE. It aims to address the limitations of the Gaussian kernel and provide a better representation of the data's local structures. 

\textbf{Isolation Kernel~\citep{zhu2021improving}: }
It measures the isolation of a data point from its neighbors. It captures the idea of how well a point can be separated from its surroundings, emphasizing the local structure preservation in the low-dimensional space. By assigning higher similarity values to points that are well-isolated, the Isolation kernel can enhance the separation of clusters in t-SNE visualizations.




\textbf{Challenges and Limitations of the Isolation Kernel: }
Despite its advantages, the Isolation kernel faces challenges and limitations that affect its performance in t-SNE. 
For example, the Isolation kernel may struggle with balancing the preservation of local and global structures.
To overcome the challenge, we propose a modified Isolation kernel that incorporates robustness measures and optimizes the balance between local and global structure preservation.

\section{MIK Theoretical Proofs}\label{sec_proof_appendix}

We provide theoretical proofs to validate the effectiveness of the modified isolation kernel in addressing the limitations of the Gaussian and isolation kernels. We prove that the MIK preserves both local and global structures by balancing the contributions of nearby and distant points. Furthermore, we demonstrate that the adaptive neighborhood weights and local density information incorporated in the kernel computation enhance the separation of clusters and improve the interpretability of t-SNE visualizations.

\begin{theorem}
\label{theorom_MIK}
    The modified isolation kernel preserves local and global structures in t-SNE visualizations.
\end{theorem}

\begin{proof}
    To prove the theorem~\ref{theorom_MIK} we need to show that the pairwise similarities computed using the modified isolation kernel (MIK) accurately represent the similarities between data points in both local and global contexts.

    Consider a pair of data points $x_i$ and $x_j$ in the high-dimensional space. The modified isolation kernel takes into account the adaptive neighborhood sizes, local densities, and distance between the points to compute the similarity between them.

    First, let's consider the case of local structure preservation. When the points $x_i$ and $x_j$ are close to each other, the adaptive neighborhood sizes will be relatively small, resulting in a larger value for $\sigma_i$ and $\sigma_j$. Additionally, if the local density of both points is high, the terms $\left(1 - \frac{n_i}{\sum_{k\neq i} n_k}\right)$ and $\left(1 - \frac{n_j}{\sum_{k\neq j} n_k}\right)$ will be closer to zero, reducing the impact of points with lower local density. Therefore, the modified isolation kernel assigns a higher similarity value between $x_i$ and $x_j$ in cases where they are part of the same local structure.

    Next, let's consider the case of global structure preservation. When the points $x_i$ and $x_j$ are far apart, the adaptive neighborhood sizes will be relatively large, resulting in smaller values for $\sigma_i$ and $\sigma_j$. This causes the exponential term in the modified isolation kernel to decrease, indicating a lower similarity between $x_i$ and $x_j$. Therefore, the MIK assigns a lower similarity value between $x_i$ and $x_j$ when they are part of different global structures.

    Based on the above analysis, we can conclude that the modified isolation kernel effectively captures both local and global structures in t-SNE visualizations, preserving the similarities between data points accurately.
\end{proof}

\begin{theorem}
    \label{theorem_Cluster}
     The modified isolation kernel enhances the separation of clusters in t-SNE visualizations.
\end{theorem}

\begin{proof}
    To prove Theorem~\ref{theorem_Cluster}, we need to demonstrate that the inclusion of adaptive neighborhood weights and local density information in the kernel computation improves the differentiation between clusters.
    Consider a set of clusters in the high-dimensional space. The modified isolation kernel takes into account the local densities of points within each cluster. When computing the similarity between two points $x_i$ and $x_j$ from different clusters, the terms $\left(1 - \frac{n_i}{\sum_{k\neq i} n_k}\right)$ and $\left(1 - \frac{n_j}{\sum_{k\neq j} n_k}\right)$ in the modified isolation kernel will be closer to zero, reducing the similarity between these points. This results in a higher dissimilarity between points from different clusters, enhancing the separation between clusters in the t-SNE visualization.

    Furthermore, the inclusion of adaptive neighborhood weights in the modified isolation kernel allows for a more flexible and adaptive influence of nearby points. The kernel considers the varying neighborhood sizes of points and adjusts the standard deviations $\sigma_i$ and $\sigma_j$ accordingly. This ensures that the similarity between points is computed based on their relative distances within their respective local neighborhoods. As a result, points within the same cluster are more likely to have higher similarities, leading to tighter clustering in the t-SNE visualization.
\end{proof}

\section{Evaluation Metrics Detail}\label{appendix_eval}
Here we discuss the evaluation metrics used for tSNE, classification, and clustering results evaluation.
\subsection{tSNE Evaluation Metrics}
\subsubsection{Neighborhood Agreement (NA)}
It measures the extent to which the local neighborhoods of data points are preserved in the low-dimensional space. It quantifies the similarity between pairwise distances in the high-dimensional space and the low-dimensional space.
Formally, let $D^H$ denote the pairwise Euclidean distances between data points in the original HD space, and $D^L$ denote the pairwise distances in the LD space. The NA is computed as :


\begin{equation}
\scriptsize
NA = 1 - \frac{2}{N (N - 1)} \sum_{i=1}^{N} \sum_{j \neq i} \left\vert \frac{d_{ij}^{H} - d_{ij}^{L}}{d_{ij}^{H} + d_{ij}^{L}} \right\vert
\end{equation}

where $N$ represents the total number of data points, $d_{ij}^H$ is the Euclidean distance between data points $i$ and $j$ in the high-dimensional space, and $d_{ij}^L$ is the Euclidean distance between their corresponding low-dimensional representations.
The Neighborhood Agreement ranges between 0 and 1, with higher values indicating better preservation of local neighborhoods.

\subsubsection{Trustworthiness (TW)}


It measures the extent to which the ranking of pairwise distances is preserved in the low-dimensional space.
Let $R_k$ denote the set of $k=100$ nearest neighbors of a data point $i$ in the high-dimensional space, and $R_k^L$ denote the corresponding set of $k=100$ nearest neighbors in the low-dimensional space. Then  Trustworthiness can be presented by:

\begin{equation}
\scriptsize
    TW = 1 - \frac{2}{N \cdot k \cdot (2N - 3k - 1)} \sum_{i=1}^{N} \sum_{j \in R_k} (R_{ij} - R_{ij}^{L})
\end{equation}

where $N$ represents total number of data points, $R_{ij}$ represents rank  $j$ in the high-dimensional space, and $R_{ij}^L$ represents the rank in the low-dimensional space. The Trustworthiness ranges between 0 and 1, with higher values indicating a more faithful preservation of pairwise distances.




\subsection{Classification Evaluation Metrics}
For classification, we use Support Vector Machine (SVM), Naive Bayes (NB), Multi-Layer Perceptron (MLP), K-Nearest Neighbour (KNN) (with $K = 5$, which decided using standard validation set approach~\citep{validationSetApproach}), Random Forest (RF), Logistic Regression (LR), and Decision Tree (DT) classifiers. 
We use average accuracy, precision, recall, weighted, and ROC area under the curve (AUC) as evaluation metrics for measuring the goodness of classification algorithms.

\subsection{Clustering Evaluation Metrics}
Following metrics are used measure the goodness of clustering:

\textbf{Silhouette Coefficient:}~\citep{rousseeuw1987silhouettes} 
Indicates how well each data point fits its assigned cluster and how distinct the clusters are from each other. Its values lie between -1 (worst) and 1 (best).


\textbf{Calinski-Harabasz Index: }~\citep{calinski1974dendrite} Also known as the Variance Ratio Criterion, is used to evaluate the compactness and separation of clusters. Higher values of the Calinski-Harabasz index indicate better-defined and more compact clusters. 

\textbf{Davies-Bouldin Index: }~\citep{davies1979cluster} 
Measures average similarity between clusters, where lower values indicate better clustering performance. It considers both the compactness and separation of clusters. 

\begin{table}[h!]
    \centering
    \resizebox{0.49\textwidth}{!}{
    \begin{tabular}{p{1.4cm}p{0.7cm}cp{0.7cm}p{0.7cm}p{0.9cm}p{4cm}}
    \toprule
    \multirow{2}{1.7cm}{Dataset} & \multirow{2}{*}{$\vert$ Seq. $\vert$} & \multirow{2}{*}{$\vert$ Classes $\vert$} & \multicolumn{3}{c}{Sequence Length}& \multirow{2}{*}{Detail} \\
    \cmidrule{4-6}
        & & & Max & Min & Mean &  \\
        \midrule
       \multirow{2}{1.1cm}{Protein Subcellular}  & \multirow{2}{*}{5959} & \multirow{2}{*}{11} & \multirow{2}{*}{3678} & \multirow{2}{*}{9} & \multirow{2}{*}{326.27} & The unaligned protein sequences having information about subcellular locations. \\
    \midrule
       \multirow{3}{2.7cm}{GISAID}  & \multirow{3}{*}{7000} & \multirow{3}{*}{22} & \multirow{3}{*}{1274} & \multirow{3}{*}{1274} & \multirow{3}{*}{1274.00} & The aligned spike sequences of the SARS-CoV-2 virus having the information about the Lineage of each sequence. \\
       \midrule
       \multirow{2}{2.7cm}{Nucleotide}  & \multirow{2}{*}{4380} & \multirow{2}{*}{7} & \multirow{2}{*}{18921} & \multirow{2}{*}{5} & \multirow{2}{*}{1263.59}& Unaligned nucleotide sequences to classify gene family to which humans belong \\
     \bottomrule
    \end{tabular}
    }
    \caption{Dataset Statistics for all three datasets. }
    \label{tbl_data_statistics}
\end{table}

\begin{figure*}[h!]
  \centering
  \subfigure[Spike2Vec (Random Init.)]{
    \includegraphics[scale=0.62]{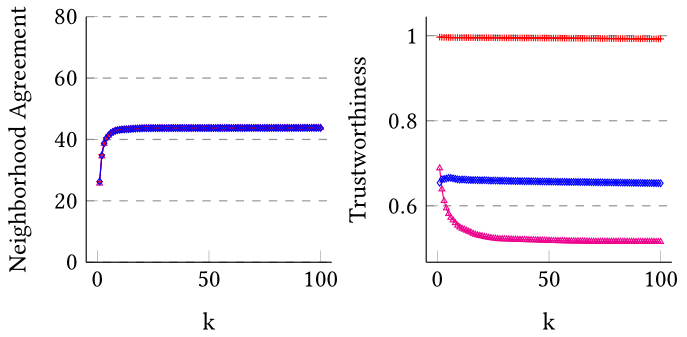}
  }%
 \subfigure[Spaced $k$-mers (Random Init.)]{
    \includegraphics[scale=0.62]{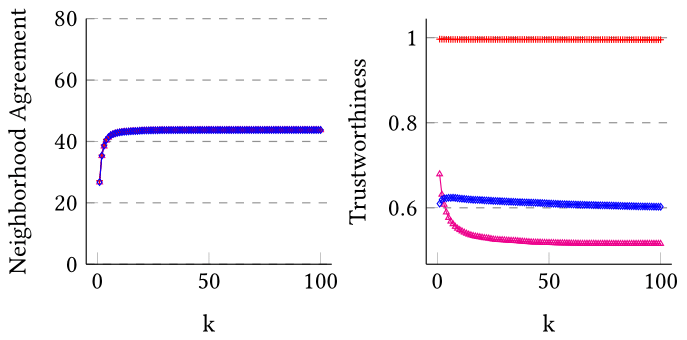}
  }%
  \subfigure[PWM2Vec (Random Init.)]{
  \includegraphics[scale=0.62]{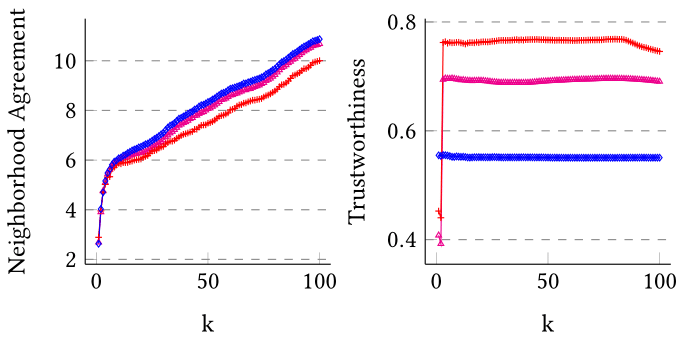}
  }%
  \\
  \subfigure[Spike2Vec (PCA Init.)]{
  \includegraphics[scale=0.62]{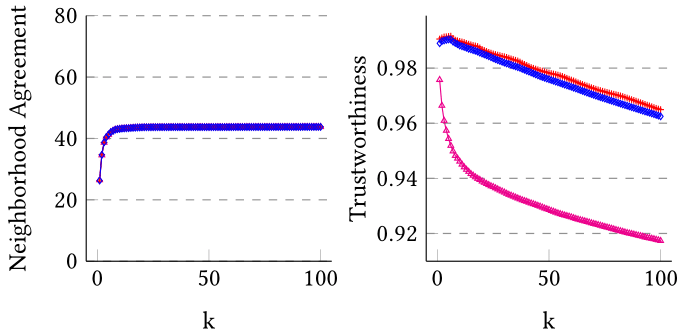}
  }%
  \subfigure[Spaced $k$-mers (PCA Init.)]{
  \includegraphics[scale=0.62]{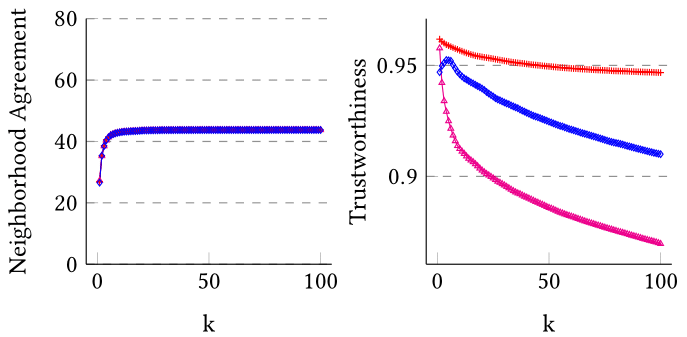}
    }%
 \subfigure[PWM2Vec (PCA Init.)]{
 \includegraphics[scale=0.62]{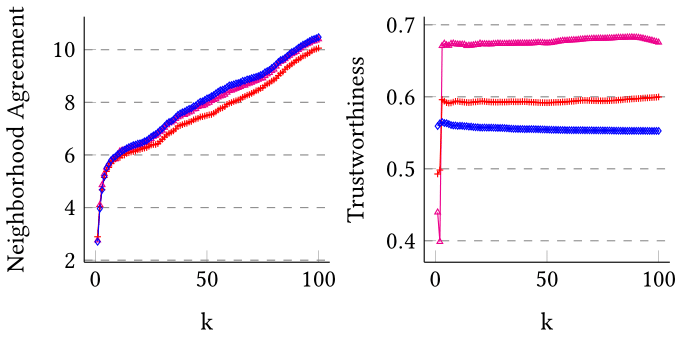}
  }%
  \\
 \subfigure[Spike2Vec (Random Walk Init.)]{\includegraphics[scale=0.62]{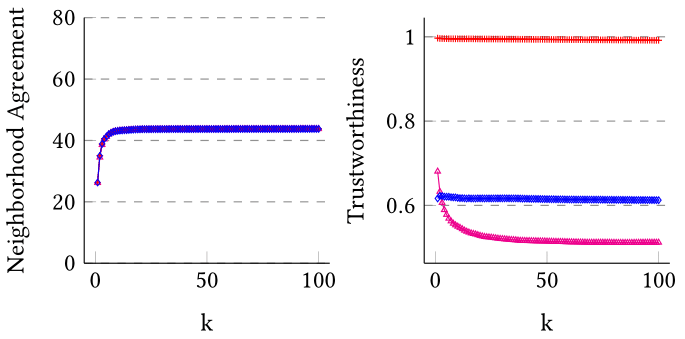}
  }%
  \subfigure[Spaced $k$-mers (Random Walk Init.)]{\includegraphics[scale=0.62]{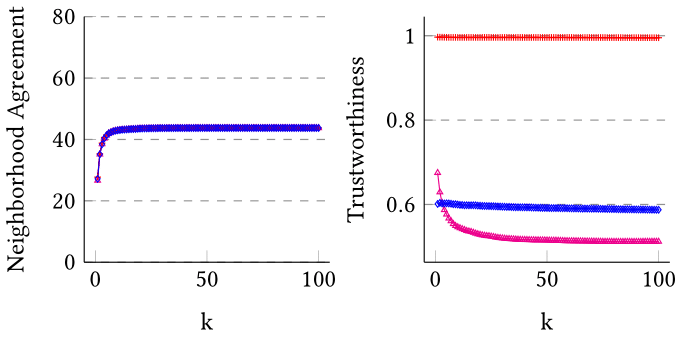}
  }%
  \subfigure[PWM2Vec (Random Walk Init.)]{
  \includegraphics[scale=0.62]{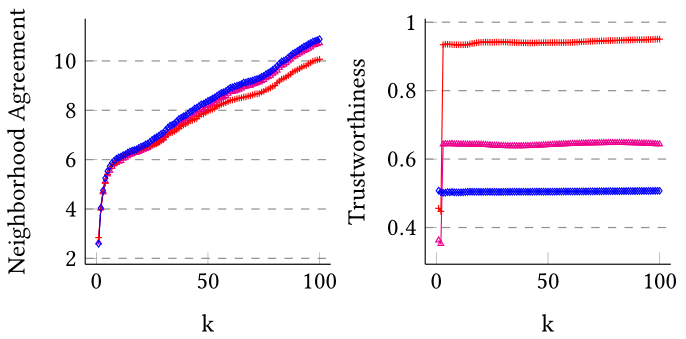}
  }%
  \\
  \includegraphics[scale=0.95]{legend.png}
  \caption{Neighborhood Agreement and Trustworthiness for tSNE-based 2D representation of \textbf{Nucleotide} Dataset using a different number of neighbors $k$ for \textbf{Spike2Vec Embedding}. These results are computed for \textbf{random initialization} of tSNE. The figure is best seen in color.} 
  \label{fig_Auc_humanDNA_spike2Vec_random_neighborhood}
\end{figure*}

\begin{figure*}[h!]
  \centering
 \subfigure[Spike2Vec (Random Init.)]{
 \includegraphics[scale=0.62]{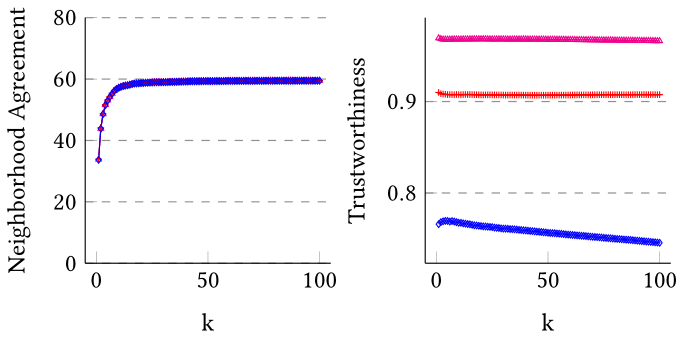}
  }%
  \subfigure[Spaced $k$-mers (Random Init.)]{
  \includegraphics[scale=0.62]{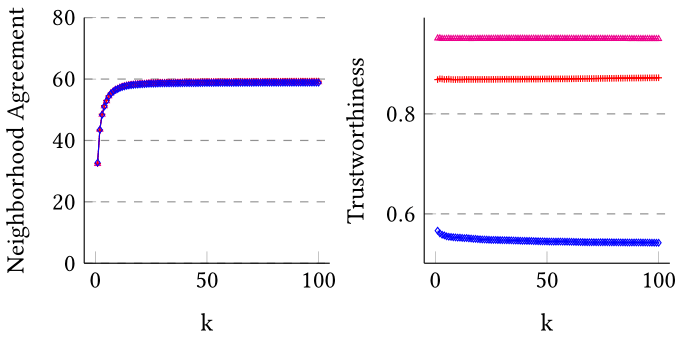}
  }%
  \subfigure[PWM2Vec (Random Init.)]{
  \includegraphics[scale=0.62]{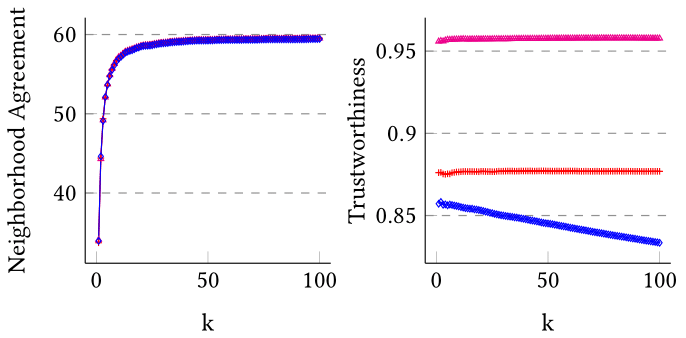}
  }%
  \\
  \subfigure[Spike2Vec (PCA Init.)]{
  \includegraphics[scale=0.62]{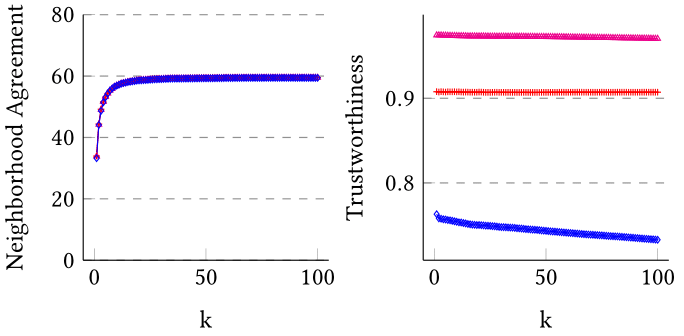}
  }%
  \subfigure[Spaced $k$-mers (PCA Init.)]{
  \includegraphics[scale=0.62]{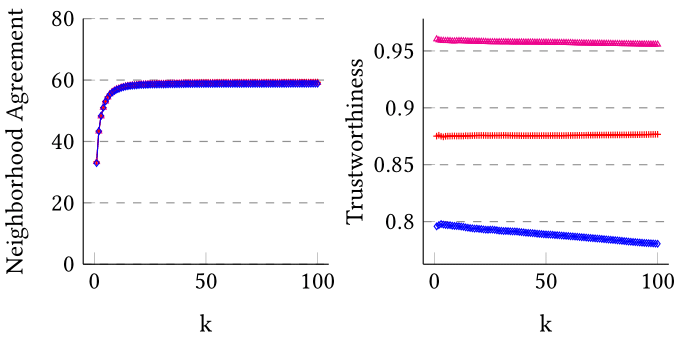}
  }%
 \subfigure[PWM2Vec (PCA Init.)]{
  \includegraphics[scale=0.62]{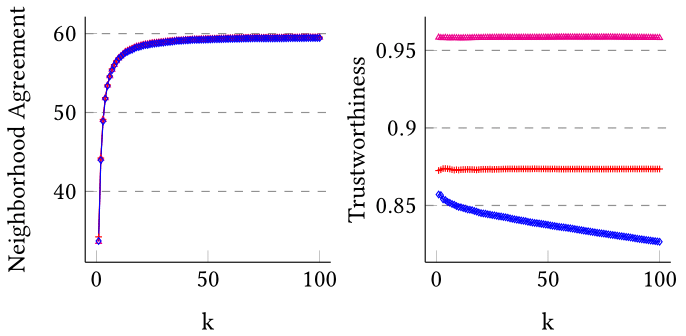}
  }%
  \\
\subfigure[Spike2Vec (Random Walk Init.)]{
  \includegraphics[scale=0.62]{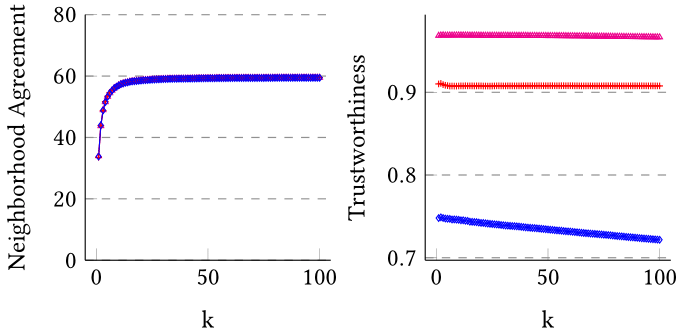}
  }%
\subfigure[Spaced $k$-mers (Random Walk Init.)]{
  \includegraphics[scale=0.62]{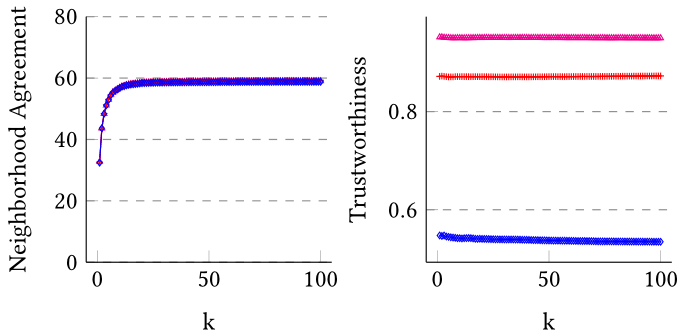}
  }%
\subfigure[PWM2Vec (Random Walk Init.)]{
  \includegraphics[scale=0.62]{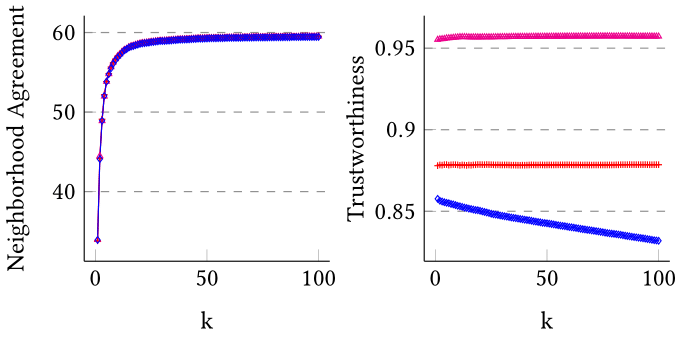}
  }%
   \\
  \includegraphics[scale=0.95]{legend.png}
  \caption{Neighborhood Agreement and Trustworthiness for tSNE-based 2D representation of \textbf{Protein Subcellular} Dataset using a different number of neighbors $k$ for \textbf{Spike2Vec Embedding}. These results are computed for the \textbf{random initialization} of tSNE. The figure is best seen in color.} 
  \label{fig_Auc_Protein_Subcellular_spike2Vec_random_neighborhood}
\end{figure*}

\begin{table}[h!]
\centering
\resizebox{0.49\textwidth}{!}{
 \begin{tabular}{@{\extracolsep{6pt}}p{0.9cm}p{0.9cm}p{0.5cm}p{0.6cm}p{0.6cm}p{0.6cm}p{0.6cm}p{0.7cm}p{0.6cm}p{0.9cm}}
    \toprule
        \multirow{2}{*}{Kernel} & \multirow{2}{*}{Embeddings} & \multirow{2}{*}{Algo.} & \multirow{2}{*}{Acc. $\uparrow$} & \multirow{2}{*}{Prec. $\uparrow$} & \multirow{2}{*}{Recall $\uparrow$} & \multirow{2}{1cm}{F1 (Weig.) $\uparrow$} & \multirow{2}{1cm}{F1 (Macro) $\uparrow$} & \multirow{2}{0.6cm}{ROC AUC $\uparrow$} & Train Time (sec.) $\downarrow$\\
        \midrule \midrule
        
        \multirow{21}{1.2cm}{Gaussian} & \multirow{7}{0.8cm}{Spike2Vec}
        & SVM & 0.4972 & 0.4974 & 0.4972 & 0.4835 & 0.3868 & 0.6564 & 11.7554 \\
 &  & NB & 0.3434 & 0.4079 & 0.3434 & 0.3471 & 0.3023 & 0.6124 & \underline{0.0344} \\
 &  & MLP & 0.4441 & 0.4204 & 0.4441 & 0.4227 & 0.2875 & 0.6111 & 18.3876 \\
 &  & KNN & \underline{0.5414} & 0.5440 & \underline{0.5414} & \underline{0.5306} & \underline{0.4329} & \underline{0.6776} & 0.1489 \\
 &  & RF & 0.5336 & \underline{0.6427} & 0.5336 & 0.4943 & 0.3324 & 0.6284 & 7.3138 \\
 &  & LR & 0.4866 & 0.4693 & 0.4866 & 0.4384 & 0.2599 & 0.6065 & 0.6357 \\
 &  & DT & 0.3960 & 0.3941 & 0.3960 & 0.3945 & 0.2809 & 0.6067 & 1.1742 \\
        \cmidrule{2-10} 
         & \multirow{7}{1.2cm}{Spaced $k$-mers}
          & SVM & 0.4771 & 0.4703 & 0.4771 & 0.4548 & 0.3312 & 0.6361 & 18.0723 \\
 &  & NB & 0.3255 & 0.3766 & 0.3255 & 0.3252 & 0.2588 & 0.6150 & \underline{0.0713} \\
 &  & MLP & 0.4290 & 0.4035 & 0.4290 & 0.4100 & 0.2661 & 0.6033 & 15.7677 \\
 &  & KNN & 0.5056 & 0.5152 & 0.5056 & 0.4954 & \underline{0.3927} & \underline{0.6584} & 0.4504 \\
 &  & RF & \underline{0.5123} &\underline{ 0.5734} & \underline{0.5123} & \underline{0.4705} & 0.3172 & 0.6205 & 8.1846 \\
 &  & LR & 0.4692 & 0.4436 & 0.4692 & 0.4180 & 0.2497 & 0.5985 & 0.5803 \\
 &  & DT & 0.3736 & 0.3704 & 0.3736 & 0.3710 & 0.2675 & 0.6052 & 0.8625 \\
 \cmidrule{2-10} 
 & \multirow{7}{0.8cm}{PWM2Vec}
          & SVM & 0.5207 & 0.5190 & 0.5207 & 0.5110 & 0.3947 & 0.6630 & 9.7228 \\
 &  & NB & 0.3876 & 0.4438 & 0.3876 & 0.3965 & 0.3206 & 0.6376 & \underline{0.0554} \\
 &  & MLP & 0.4640 & 0.4424 & 0.4640 & 0.4487 & 0.2990 & 0.6196 & 9.8790 \\
 &  & KNN & \underline{0.5634} & 0.5671 & \underline{0.5634} & \underline{0.5550} & \underline{0.4567} & \underline{0.6950} & 0.1219 \\
 &  & RF & 0.5252 & \underline{0.6064} & 0.5252 & 0.4835 & 0.3176 & 0.6215 & 6.8668 \\
 &  & LR & 0.5085 & 0.4830 & 0.5085 & 0.4654 & 0.2709 & 0.6106 & 0.4873 \\
 &  & DT & 0.3668 & 0.3702 & 0.3668 & 0.3677 & 0.2597 & 0.5978 & 0.6648 \\

 \midrule
         \multirow{21}{1.2cm}{Isolation} & \multirow{7}{0.8cm}{Spike2Vec}
        & SVM & 0.6493 & 0.6550 & 0.6493 & 0.6487 & 0.4977 & 0.7326 & 9.7372 \\
 &  & NB & 0.2701 & 0.3615 & 0.2701 & 0.2285 & 0.2426 & 0.6714 & \underline{0.0455} \\
 &  & MLP & 0.7796 & 0.7727 & 0.7796 & 0.7752 & 0.6105 & 0.7917 & 11.8214 \\
 &  & KNN & \underline{0.8546} & \underline{0.8595} & \underline{0.8546} & \underline{0.8522} & \underline{0.7464} & \underline{0.8622} & 0.1707 \\
 &  & RF & 0.8345 & 0.8538 & 0.8345 & 0.8069 & 0.5968 & 0.7712 & 5.9505 \\
 &  & LR & 0.2450 & 0.0600 & 0.2450 & 0.0964 & 0.0358 & 0.5000 & 0.2048 \\
 &  & DT & 0.6913 & 0.6973 & 0.6913 & 0.6919 & 0.5203 & 0.7430 & 0.6885 \\
         \cmidrule{2-10} 
         & \multirow{7}{1.2cm}{Spaced $k$-mers}
        & SVM & 0.4961 & 0.5010 & 0.4961 & 0.4833 & 0.3431 & 0.6408 & 11.4647 \\
 &  & NB & 0.1186 & 0.2258 & 0.1186 & 0.1002 & 0.0980 & 0.5592 & \underline{0.0430} \\
 &  & MLP & \underline{0.7601} & \underline{0.7432} & \underline{0.7601} & \underline{0.7493} & 0.5492 & \underline{0.7578} & 12.5552 \\
 &  & KNN & 0.7002 & 0.6989 & 0.7002 & 0.6938 & \underline{0.5566} & 0.7547 & 0.1021 \\
 &  & RF & 0.7226 & 0.7320 & 0.7226 & 0.6911 & 0.4678 & 0.7088 & 5.2552 \\
 &  & LR & 0.2438 & 0.0595 & 0.2438 & 0.0956 & 0.0356 & 0.5000 & 0.2202 \\
 &  & DT & 0.5419 & 0.5535 & 0.5419 & 0.5470 & 0.3836 & 0.6685 & 0.5416 \\
         \cmidrule{2-10} 
 &  \multirow{7}{0.8cm}{PWM2Vec}
          & SVM & 0.6081 & 0.6197 & 0.6081 & 0.6052 & 0.4713 & 0.7177 & 7.1583 \\
 &  & NB & 0.1912 & 0.3872 & 0.1912 & 0.1836 & 0.1932 & 0.6221 & \underline{0.0425} \\
 &  & MLP & 0.7496 & 0.7270 & 0.7496 & 0.7353 & 0.5332 & 0.7507 & 7.6877 \\
 &  & KNN & \underline{0.8150} & \underline{0.8135} & \underline{0.8150} & \underline{0.8074} & \underline{0.6564} & \underline{0.8128} & 0.0791 \\
 &  & RF & 0.7686 & 0.7709 & 0.7686 & 0.7281 & 0.4787 & 0.7222 & 5.3885 \\
 &  & LR & 0.2293 & 0.0526 & 0.2293 & 0.0856 & 0.0339 & 0.5000 & 0.2301 \\
 &  & DT & 0.5888 & 0.5942 & 0.5888 & 0.5909 & 0.4167 & 0.6884 & 0.5060 \\
         \midrule
    \multirow{21}{1.2cm}{Modified Isolation (Ours)} & \multirow{7}{0.8cm}{Spike2Vec}
 & SVM & 0.7864 & 0.7932 & 0.7864 & 0.7875 & 0.6742 & 0.8296 & 6.4243 \\
 &  & NB & 0.4871 & 0.5988 & 0.4871 & 0.4704 & 0.4424 & 0.7628 & \underline{0.0631} \\
 &  & MLP & \underline{0.8619} & \underline{0.8592} & \underline{0.8619} & \underline{0.8592} & \underline{0.7483} & \underline{0.8605} & 10.8744 \\
 &  & KNN & 0.8462 & 0.8515 & 0.8462 & 0.8403 & 0.7205 & 0.8387 & 0.1028 \\
 &  & RF & 0.8451 & 0.8429 & 0.8451 & 0.8157 & 0.5884 & 0.7774 & 8.0731 \\
 &  & LR & 0.2360 & 0.1810 & 0.2360 & 0.0927 & 0.0360 & 0.5003 & 0.3017 \\
 &  & DT & 0.6890 & 0.6870 & 0.6890 & 0.6877 & 0.5268 & 0.7450 & 1.2233 \\
        \cmidrule{2-10} 
         & \multirow{7}{1.2cm}{Spaced $k$-mers}
        & SVM & 0.8697 & 0.8767 & 0.8697 & 0.8712 & 0.7327 & 0.8663 & 1.1591 \\
 &  & NB & 0.4010 & 0.5894 & 0.4010 & 0.3824 & 0.3661 & 0.7481 & \underline{0.0364} \\
 &  & MLP & 0.8937 & \textbf{\underline{0.8971}} & 0.8937 & \textbf{\underline{0.8949}} & \textbf{\underline{0.7847}} & \textbf{\underline{0.8907}} & 10.0736 \\
 &  & KNN & 0.6119 & 0.6231 & 0.6119 & 0.5914 & 0.4010 & 0.6817 & 0.1299 \\
 &  & RF & \textbf{\underline{0.8982}} & 0.8947 & \textbf{\underline{0.8982}} & 0.8867 & 0.7125 & 0.8443 & 6.3723 \\
 &  & LR & 0.2204 & 0.2584 & 0.2204 & 0.0811 & 0.0337 & 0.5003 & 0.2525 \\
 &  & DT & 0.7265 & 0.7377 & 0.7265 & 0.7301 & 0.5701 & 0.7793 & 1.0005 \\
\cmidrule{2-10} 
     & \multirow{7}{1.2cm}{PWM2Vec}
          & SVM & 0.7714 & 0.7796 & 0.7714 & 0.7727 & 0.6648 & 0.8270 & 5.0938 \\
 &  & NB & 0.4829 & 0.5515 & 0.4829 & 0.4681 & 0.4637 & 0.7541 & \textbf{\underline{0.0315}} \\
 &  & MLP & 0.8416 & 0.8430 & 0.8416 & 0.8416 & 0.6866 & 0.8364 & 6.3770 \\
 &  & KNN & \underline{0.8610} & \underline{0.8666} & \underline{0.8610} & \underline{0.8585} & \underline{0.7441} & \underline{0.8554} & 0.0800 \\
 &  & RF & 0.8323 & 0.8489 & 0.8323 & 0.8039 & 0.5878 & 0.7712 & 6.5016 \\
 &  & LR & 0.2320 & 0.2066 & 0.2320 & 0.0894 & 0.0353 & 0.5003 & 0.2440 \\
 &  & DT & 0.6813 & 0.6862 & 0.6813 & 0.6830 & 0.5164 & 0.7440 & 0.6294 \\
         \bottomrule
         \end{tabular}
}
 \caption{Classification results (averaged over $5$ runs) on \textbf{Protein Subcellular} dataset. The best values are underlined for each embedding method, and overall best values are shown in bold.}
    \label{tbl_results_classification_protein_subcellular}
\end{table}

\begin{table}[h!]
\centering
\resizebox{0.49\textwidth}{!}{
  \begin{tabular}{@{\extracolsep{6pt}}p{0.9cm}p{0.9cm}p{0.5cm}p{0.6cm}p{0.6cm}p{0.6cm}p{0.6cm}p{0.7cm}p{0.6cm}p{0.9cm}}
    \toprule
        \multirow{2}{*}{Kernel} & \multirow{2}{*}{Embeddings} & \multirow{2}{*}{Algo.} & \multirow{2}{*}{Acc. $\uparrow$} & \multirow{2}{*}{Prec. $\uparrow$} & \multirow{2}{*}{Recall $\uparrow$} & \multirow{2}{1cm}{F1 (Weig.) $\uparrow$} & \multirow{2}{1cm}{F1 (Macro) $\uparrow$} & \multirow{2}{0.6cm}{ROC AUC $\uparrow$} & Train Time (sec.) $\downarrow$\\
        \midrule \midrule
        \multirow{21}{1.2cm}{Gaussian} & \multirow{7}{0.8cm}{Spike2Vec}
          & SVM & 0.7397 & 0.6998 & 0.7397 & 0.7009 & 0.3060 & 0.6541 & 3.4177 \\
 &  & NB & 0.1564 & 0.6631 & 0.1564 & 0.2226 & 0.1312 & 0.5745 & \textbf{\underline{0.0866}} \\
 &  & MLP & 0.7696 & 0.7308 & 0.7696 & 0.7388 & 0.4025 & 0.7111 & 8.2361 \\
 &  & KNN & 0.7678 & 0.7725 & 0.7678 & 0.7634 & 0.5223 & 0.7560 & 0.1034 \\
 &  & RF & \textbf{\underline{0.7872}} & \underline{0.7809} & \textbf{\underline{0.7872}} & \textbf{\underline{0.7762}} & \underline{0.5342} & \textbf{\underline{0.7637}} & 2.8961 \\
 &  & LR & 0.7174 & 0.6514 & 0.7174 & 0.6686 & 0.2447 & 0.6226 & 1.2611 \\
 &  & DT & 0.7756 & 0.7702 & 0.7756 & 0.7668 & 0.5106 & 0.7550 & 0.2956 \\
        \cmidrule{2-10} 
        & \multirow{7}{1.2cm}{Spaced $k$-mers}
         & SVM & 0.7214 & 0.6975 & 0.7214 & 0.6894 & 0.4974 & 0.7145 & 8.3873 \\
          &  & NB & 0.5419 & \underline{0.7841} & 0.5419 & 0.5767 & 0.3988 & 0.7456 & \underline{0.1246} \\
 &  & MLP & 0.7560 & 0.7368 & 0.7560 & 0.7212 & 0.5436 & 0.7429 & 5.9533 \\
 &  & KNN & 0.7265 & 0.7222 & 0.7265 & 0.7138 & 0.5382 & 0.7474 & 0.1803 \\
 &  & RF & \underline{0.7575} & 0.7410 & \underline{0.7575} & 0.7316 & \textbf{\underline{0.5595}} & 0.7547 & 3.2786 \\
 &  & LR & 0.7456 & 0.7065 & 0.7456 & 0.7007 & 0.5087 & 0.7267 & 1.3746 \\
 &  & DT & 0.7544 & 0.7400 & 0.7544 & \underline{0.7318} & 0.5594 & \underline{0.7571} & 0.3411 \\
        \cmidrule{2-10} 
        & \multirow{7}{0.8cm}{PWM2Vec}
         & SVM & 0.7261 & 0.6742 & 0.7261 & 0.6699 & 0.4792 & 0.7017 & 6.7129 \\
          &  & NB & 0.5492 & \textbf{\underline{0.7868}} & 0.5492 & 0.5926 & 0.3597 & 0.7250 & \underline{0.0912} \\
 &  & MLP & 0.7425 & 0.7253 & 0.7425 & 0.7054 & 0.5175 & 0.7275 & 5.4593 \\
 &  & KNN & 0.7224 & 0.7295 & 0.7224 & 0.7125 & 0.5254 & 0.7367 & 0.1544 \\
 &  & RF & \underline{0.7549} & 0.7363 & \underline{0.7549} & \underline{0.7251} & \underline{0.5383} & \underline{0.7395} & 3.2420 \\
 &  & LR & 0.7316 & 0.6960 & 0.7316 & 0.6847 & 0.4884 & 0.7108 & 1.4509 \\
 &  & DT & 0.7490 & 0.7282 & 0.7490 & 0.7221 & 0.5300 & 0.7387 & 0.2997 \\
        \midrule
        \multirow{21}{1.2cm}{Isolation} & \multirow{7}{0.8cm}{Spike2Vec}
        & SVM & 0.4703 & 0.3027 & 0.4703 & 0.3429 & 0.0469 & 0.5061 & 40.9601 \\
 &  & NB & 0.0173 & 0.2582 & 0.0173 & 0.0156 & 0.0222 & 0.5077 & \underline{0.1303} \\
 &  & MLP & \underline{0.6523} & 0.6052 & \underline{0.6523} & \underline{0.6197} & 0.1476 & 0.5738 & 15.3755 \\
 &  & KNN & 0.5497 & 0.5099 & 0.5497 & 0.5258 & 0.1067 & 0.5433 & 0.1546 \\
 &  & RF & 0.6305 & 0.6111 & 0.6305 & 0.6183 & \underline{0.1726} & 0.5755 & 6.5307 \\
 &  & LR & 0.4770 & 0.2276 & 0.4770 & 0.3081 & 0.0294 & 0.5000 & 0.8537 \\
 &  & DT & 0.6080 & \underline{0.6144} & 0.6080 & 0.6103 & 0.1717 & \underline{0.5773} & 0.7293 \\
         \cmidrule{2-10} 
        & \multirow{7}{1.2cm}{Spaced $k$-mers}
        & SVM & 0.4972 & 0.3674 & 0.4972 & 0.3814 & 0.1007 & 0.5331 & 11.9771 \\
 &  & NB & 0.0276 & 0.2183 & 0.0276 & 0.0271 & 0.0314 & 0.5203 & \underline{0.1053} \\
 &  & MLP & 0.6499 & 0.6187 & 0.6499 & 0.6276 & 0.1884 & 0.5922 & 14.1174 \\
 &  & KNN & 0.6490 & 0.6460 & 0.6490 & 0.6356 & 0.2063 & 0.5892 & 0.1331 \\
 &  & RF & \underline{0.7029} & \underline{0.6777} & \underline{0.7029} & \underline{0.6799} & \underline{0.3469} & \underline{0.6488} & 6.2961 \\
 &  & LR & 0.4883 & 0.2385 & 0.4883 & 0.3205 & 0.0298 & 0.5000 & 0.6894 \\
 &  & DT & 0.6379 & 0.6457 & 0.6379 & 0.6404 & 0.2828 & 0.6329 & 0.8870 \\
         \cmidrule{2-10} 
          & \multirow{7}{0.8cm}{PWM2Vec}
    & SVM & 0.4948 & 0.3752 & 0.4948 & 0.3839 & 0.0905 & 0.5261 & 9.2996 \\
 &  & NB & 0.0259 & 0.2166 & 0.0259 & 0.0243 & 0.0276 & 0.5176 & \underline{0.1129} \\
 &  & MLP & 0.6425 & 0.6034 & 0.6425 & 0.6173 & 0.1796 & 0.5892 & 15.3638 \\
 &  & KNN & 0.6178 & 0.5958 & 0.6178 & 0.5967 & 0.1777 & 0.5743 & 0.1346 \\
 &  & RF & \underline{0.6870} & \underline{0.6551} & \underline{0.6870} & \underline{0.6594} & \underline{0.3257} & \underline{0.6370} & 6.1128 \\
 &  & LR & 0.4801 & 0.2305 & 0.4801 & 0.3114 & 0.0295 & 0.5000 & 0.7037 \\
 &  & DT & 0.6122 & 0.6189 & 0.6122 & 0.6146 & 0.2664 & 0.6243 & 0.9140 \\
         \midrule
        \multirow{21}{1.2cm}{Modified Isolation (Ours)} & \multirow{7}{0.8cm}{Spike2Vec}
        & SVM & 0.5554 & 0.4731 & 0.5554 & 0.5008 & 0.1086 & 0.5400 & 19.8189 \\
 &  & NB & 0.0379 & 0.2890 & 0.0379 & 0.0389 & 0.0209 & 0.5117 & \underline{0.0954} \\
 &  & MLP & \underline{0.6188} & 0.5667 & \underline{0.6188} & \underline{0.5874} & 0.1364 & 0.5671 & 10.4927 \\
 &  & KNN & 0.5922 & 0.5590 & 0.5922 & 0.5721 & 0.1281 & 0.5548 & 0.1105 \\
 &  & RF & 0.6073 & 0.5677 & 0.6073 & 0.5826 & \underline{0.1648} & 0.5686 & 6.8307 \\
 &  & LR & 0.4823 & 0.2327 & 0.4823 & 0.3139 & 0.0296 & 0.5000 & 0.6321 \\
 &  & DT & 0.5728 & \underline{0.5803} & 0.5728 & 0.5757 & 0.1604 & \underline{0.5700} & 0.7337 \\
         \cmidrule{2-10} 
        & \multirow{7}{1.2cm}{Spaced $k$-mers}
          & SVM & 0.6836 & 0.6771 & 0.6836 & 0.6767 & 0.2819 & 0.6304 & 6.7199 \\
 &  & NB & 0.5926 & 0.6087 & 0.5926 & 0.5934 & 0.1999 & 0.5942 & \underline{0.1282} \\
 &  & MLP & 0.6803 & 0.6619 & 0.6803 & 0.6673 & 0.2464 & 0.6176 & 17.7629 \\
 &  & KNN & 0.6675 & 0.6695 & 0.6675 & 0.6612 & 0.2309 & 0.6039 & 0.1625 \\
 &  & RF & \underline{0.6931} & \underline{0.6818} & \underline{0.6931} & \underline{0.6847} & \underline{0.2949} & \underline{0.6334} & 11.3797 \\
 &  & LR & 0.4768 & 0.2274 & 0.4768 & 0.3079 & 0.0294 & 0.5000 & 0.7848 \\
 &  & DT & 0.6636 & 0.6653 & 0.6636 & 0.6635 & 0.2776 & 0.6294 & 1.3944 \\
         \cmidrule{2-10} 
    & \multirow{7}{0.8cm}{PWM2Vec}
    & SVM & 0.6820 & 0.6764 & 0.6820 & 0.6762 & 0.3117 & 0.6446 & 12.7645 \\
 &  & NB & 0.5952 & 0.6176 & 0.5952 & 0.5962 & 0.2017 & 0.6008 & \underline{0.1153} \\
 &  & MLP & 0.6741 & 0.6552 & 0.6741 & 0.6584 & 0.2493 & 0.6194 & 13.1291 \\
 &  & KNN & 0.6559 & 0.6493 & 0.6559 & 0.6453 & 0.2222 & 0.5966 & 0.1637 \\
 &  & RF & \underline{0.7103} & \underline{0.6955} & \underline{0.7103} & \underline{0.6983} & \underline{0.3612} & \underline{0.6619} & 12.8492 \\
 &  & LR & 0.4748 & 0.2255 & 0.4748 & 0.3057 & 0.0293 & 0.5000 & 0.6317 \\
 &  & DT & 0.6709 & 0.6752 & 0.6709 & 0.6717 & 0.3168 & 0.6514 & 1.6763 \\
         \bottomrule
         \end{tabular}
}
 \caption{Classification results (averaged over $5$ runs) on \textbf{GISAID} dataset for different evaluation metrics. The best values are underlined for each embedding method and overall best values are shown in bold.}
    \label{tbl_results_classification_GISAID}
\end{table}

\begin{table}[h!]
\centering
\resizebox{0.49\textwidth}{!}{
 \begin{tabular}{@{\extracolsep{6pt}}p{0.9cm}p{0.9cm}p{0.5cm}p{0.6cm}p{0.6cm}p{0.6cm}p{0.6cm}p{0.7cm}p{0.6cm}p{0.9cm}}
    \toprule
        \multirow{2}{*}{Kernel} & \multirow{2}{*}{Embeddings} & \multirow{2}{*}{Algo.} & \multirow{2}{*}{Acc. $\uparrow$} & \multirow{2}{*}{Prec. $\uparrow$} & \multirow{2}{*}{Recall $\uparrow$} & \multirow{2}{1cm}{F1 (Weig.) $\uparrow$} & \multirow{2}{1cm}{F1 (Macro) $\uparrow$} & \multirow{2}{0.6cm}{ROC AUC $\uparrow$} & Train Time (sec.) $\downarrow$\\
        \midrule \midrule
        \multirow{21}{1.2cm}{Gaussian} & \multirow{7}{0.8cm}{Spike2Vec}
    & SVM & 0.3659 & 0.7266 & 0.3659 & 0.2669 & 0.2108 & 0.5458 & 0.9052 \\
 &  & NB & 0.1251 & 0.6027 & 0.1251 & 0.1264 & 0.1316 & 0.5383 & \underline{0.0134} \\
 &  & MLP & 0.3878 & \underline{0.7679} & 0.3878 & 0.2979 & 0.2411 & 0.5575 & 3.3784 \\
 &  & KNN & 0.2650 & 0.3899 & 0.2650 & 0.2477 & 0.2254 & 0.5560 & 0.0597 \\
 &  & RF & 0.4265 & 0.6787 & 0.4265 & 0.3645 & 0.3217 & 0.5891 & 2.0797 \\
 &  & LR & 0.3604 & 0.7463 & 0.3604 & 0.2501 & 0.1868 & 0.5382 & 0.1351 \\
 &  & DT & \underline{0.4274} & 0.6628 & \underline{0.4274} & \underline{0.3663} & \underline{0.3246} & \underline{0.5907} & 0.2892 \\
        \cmidrule{2-10} 
       & \multirow{7}{1.2cm}{Spaced $k$-mers}
        & SVM & 0.3688 & 0.6767 & 0.3688 & 0.2707 & 0.2116 & 0.5447 & 1.0749 \\
 &  & NB & 0.1317 & 0.7136 & 0.1317 & 0.1384 & 0.1435 & 0.5411 & \underline{0.0125} \\
 &  & MLP & 0.3857 & 0.7391 & 0.3857 & 0.2926 & 0.2315 & 0.5533 & 3.1893 \\
 &  & KNN & 0.2697 & 0.3876 & 0.2697 & 0.2560 & 0.2308 & 0.5560 & 0.0560 \\
 &  & RF & \underline{0.4295} & 0.6904 & \underline{0.4295} & \underline{0.3664} & 0.3188 & 0.5873 & 2.0117 \\
 &  & LR & 0.3612 & \underline{0.7508} & 0.3612 & 0.2478 & 0.1825 & 0.5360 & 0.1341 \\
 &  & DT & 0.4285 & 0.6625 & 0.4285 & 0.3663 & \underline{0.3198} & \underline{0.5880} & 0.2364 \\
        \cmidrule{2-10} 
        & \multirow{7}{0.8cm}{PWM2Vec}
          & SVM & 0.2970 & 0.2900 & 0.2970 & 0.1616 & 0.0953 & 0.5055 & 1.8065 \\
         & & NB & 0.0860 & \underline{0.3549} & 0.0860 & 0.0550 & 0.0610 & 0.5067 & \underline{0.0114} \\
         & & MLP & 0.3015 & 0.3142 & 0.3015 & 0.1656 & 0.0981 & 0.5073 & 3.2468 \\
         & & KNN & 0.1860 & 0.1940 & 0.1860 & \underline{0.1750} & \underline{0.1362} & 0.5054 & 0.1020 \\
         & & RF & \underline{0.3023} & 0.2973 & \underline{0.3023} & 0.1680 & 0.1004 & 0.5079 & 1.0036 \\
         & & LR & \underline{0.3023} & 0.2811 & \underline{0.3023} & 0.1524 & 0.0802 & 0.5039 & 0.1832 \\
         & & DT & 0.3014 & 0.2964 & 0.3014 & 0.1694 & 0.1034 & \underline{0.5085} & 0.0582 \\
        \midrule
         \multirow{21}{1.2cm}{Isolation} & \multirow{7}{0.8cm}{Spike2Vec}
    & SVM & 0.3213 & 0.3221 & 0.3213 & 0.3079 & 0.2588 & 0.5728 & 3.8166 \\
 &  & NB & 0.2323 & 0.4337 & 0.2323 & 0.1836 & 0.2043 & 0.5560 & \underline{0.0123} \\
 &  & MLP & 0.5275 & 0.5203 & 0.5275 & 0.5192 & 0.4616 & 0.6858 & 7.3494 \\
 &  & KNN & 0.5283 & 0.5340 & 0.5283 & 0.5275 & 0.4976 & 0.7093 & 0.0585 \\
 &  & RF & \underline{0.7469} & \underline{0.7539} & \underline{0.7469} & \underline{0.7452} & \underline{0.7345} & \underline{0.8314} & 3.4126 \\
 &  & LR & 0.3105 & 0.0965 & 0.3105 & 0.1472 & 0.0677 & 0.5000 & 0.0795 \\
 &  & DT & 0.6151 & 0.6161 & 0.6151 & 0.6150 & 0.5866 & 0.7605 & 0.3282 \\
         \cmidrule{2-10} 
        & \multirow{7}{1.2cm}{Spaced $k$-mers}
    & SVM & 0.3125 & 0.3066 & 0.3125 & 0.3025 & 0.2505 & 0.5674 & 3.8297 \\
 &  & NB & 0.2248 & 0.4360 & 0.2248 & 0.1771 & 0.1969 & 0.5509 & \underline{0.0106} \\
 &  & MLP & 0.5247 & 0.5219 & 0.5247 & 0.5159 & 0.4674 & 0.6865 & 7.2932 \\
 &  & KNN & 0.5454 & 0.5534 & 0.5454 & 0.5454 & 0.5164 & 0.7175 & 0.0592 \\
 &  & RF & \underline{0.7461} & \underline{0.7652} & \underline{0.7461} & \underline{0.7443} & \underline{0.7355} & \underline{0.8253} & 3.5677 \\
 &  & LR & 0.3056 & 0.0935 & 0.3056 & 0.1432 & 0.0669 & 0.5000 & 0.0590 \\
 &  & DT & 0.6009 & 0.6025 & 0.6009 & 0.6008 & 0.5709 & 0.7504 & 0.3547 \\
         \cmidrule{2-10} 
     & \multirow{7}{0.8cm}{PWM2Vec}
    & SVM & 0.3399 & 0.3194 & 0.3399 & 0.2769 & 0.2050 & 0.5447 & 2.2195 \\
 &  & NB & 0.0989 & 0.3679 & 0.0989 & 0.0643 & 0.0573 & 0.5085 & \underline{0.0128} \\
 &  & MLP & 0.3726 & 0.3455 & 0.3726 & 0.3495 & 0.2835 & 0.5881 & 5.1376 \\
 &  & KNN & \underline{0.5152} & \underline{0.5205} & \underline{0.5152} & \underline{0.5157} & \underline{0.4888} & \underline{0.7041} & 0.0561 \\
 &  & RF & 0.5002 & 0.4986 & 0.5002 & 0.4885 & 0.4427 & 0.6706 & 1.5011 \\
 &  & LR & 0.3102 & 0.0964 & 0.3102 & 0.1470 & 0.0676 & 0.5000 & 0.0897 \\
 &  & DT & 0.4820 & 0.4788 & 0.4820 & 0.4759 & 0.4356 & 0.6698 & 0.1227 \\
         \midrule

        \multirow{21}{1.2cm}{Modified Isolation (Ours)} & \multirow{7}{0.8cm}{Spike2Vec}
    & SVM & 0.5670 & 0.5666 & 0.5670 & 0.5621 & 0.5271 & 0.7245 & 2.6061 \\
 &  & NB & 0.2562 & 0.3668 & 0.2562 & 0.2364 & 0.2427 & 0.5653 & \underline{0.0107} \\
 &  & MLP & 0.6196 & 0.6175 & 0.6196 & 0.6160 & 0.5723 & 0.7501 & 6.6650 \\
 &  & KNN & 0.5183 & 0.5246 & 0.5183 & 0.5172 & 0.4861 & 0.7038 & 0.0569 \\
 &  & RF & \underline{0.7470} & \underline{0.7521} & \underline{0.7470} & \underline{0.7443} & \underline{0.7326} & \underline{0.8308} & 4.1174 \\
 &  & LR & 0.3142 & 0.0989 & 0.3142 & 0.1504 & 0.0683 & 0.5000 & 0.0874 \\
 &  & DT & 0.6219 & 0.6243 & 0.6219 & 0.6224 & 0.5953 & 0.7658 & 0.4265 \\
         \cmidrule{2-10} 
       & \multirow{7}{1.2cm}{Spaced $k$-mers}
    & SVM & 0.5798 & 0.5762 & 0.5798 & 0.5728 & 0.5358 & 0.7273 & 2.6021 \\
 &  & NB & 0.2604 & 0.3666 & 0.2604 & 0.2363 & 0.2463 & 0.5700 & \textbf{\underline{0.0080}} \\
 &  & MLP & 0.6207 & 0.6208 & 0.6207 & 0.6178 & 0.5734 & 0.7498 & 7.1481 \\
 &  & KNN & 0.5050 & 0.5093 & 0.5050 & 0.5041 & 0.4755 & 0.6954 & 0.0611 \\
 &  & RF & \textbf{\underline{0.7481}} & \textbf{\underline{0.7689}} & \textbf{\underline{0.7481}} & \textbf{\underline{0.7455}} & \textbf{\underline{0.7366}} & \textbf{\underline{0.8316}} & 4.1538 \\
 &  & LR & 0.3096 & 0.1191 & 0.3096 & 0.1466 & 0.0679 & 0.5001 & 0.0931 \\
 &  & DT & 0.6237 & 0.6257 & 0.6237 & 0.6237 & 0.5959 & 0.7644 & 0.3863 \\
         \cmidrule{2-10} 
       & \multirow{7}{0.8cm}{PWM2Vec}
          & SVM & 0.4696 & 0.4688 & 0.4696 & 0.4497 & 0.4029 & 0.6462 & 8.9142 \\
 &  & NB & 0.1980 & 0.3166 & 0.1980 & 0.1692 & 0.1793 & 0.5507 & \underline{0.0129} \\
 &  & MLP & 0.5300 & 0.5296 & 0.5300 & 0.5183 & 0.4843 & 0.6935 & 5.4553 \\
 &  & KNN & 0.5359 & 0.5431 & 0.5359 & 0.5369 & 0.5097 & 0.7155 & 0.0590 \\
 &  & RF & \underline{0.5878} & \underline{0.5903} & \underline{0.5878} & \underline{0.5840} & \underline{0.5601} & \underline{0.7363} & 1.7506 \\
 &  & LR & 0.3119 & 0.0974 & 0.3119 & 0.1484 & 0.0679 & 0.5000 & 0.0738 \\
 &  & DT & 0.5791 & 0.5811 & 0.5791 & 0.5764 & 0.5487 & 0.7332 & 0.1447 \\
         \bottomrule
         \end{tabular}
}
 \caption{Classification results (averaged over $5$ runs) on \textbf{Nucleotide} dataset for different evaluation metrics. The best values are underlined for each embedding method and overall best values are shown in bold.}
    \label{tbl_results_classification_Human_DNA}
\end{table}

\section{Statistical Significance Of Classification Results}\label{sec_stats_classi_appendix}
To evaluate the stability and relevance of classification results, we computed p-values using the average and std. of all evaluation metrics (for $5$ runs of experiments) for all three datasets. We noted that the p-values for all pairwise comparisons were $<0.05$ for all but one evaluation metric (training runtime), showing the statistical significance of the results. For all evaluation metrics (other than training runtime), we observed standard deviations $<0.001$ for $5$ runs of experiments with random splits. For training runtime, we observed that there is a comparatively higher variation in the values, hence the p-values were sometimes $>0.05$. This is due to the fact the training time can be affected by many factors, such as processor performance and the number of jobs at any given time.

\begin{table}[h!]
\centering
\resizebox{0.49\textwidth}{!}{
 \begin{tabular}{@{\extracolsep{6pt}}p{1.2cm}llp{1.6cm}p{1.4cm}p{1.4cm}}
    \toprule
        \multirow{1}{*}{Kernel} & \multirow{1}{*}{Embeddings} & \multirow{1}{*}{Algo.} & \multirow{1}{*}{Silhouette $\uparrow$} & \multirow{1}{*}{Calinski $\uparrow$} & \multirow{1}{*}{Davies $\downarrow$}\\
        \midrule \midrule
         
        \multirow{9}{1.2cm}{Gaussian} &
        \multirow{3}{0.8cm}{Spike2Vec}
         & K-means & 0.101796 & 360.804549 & 3.622216 \\
         &  & Agglomerative & 0.078076 & 322.784851 & 3.585005 \\
         &  & K-Modes & -0.359683 & 1.646458 & 1.119053 \\
        \cmidrule{2-6} 
        & \multirow{3}{1.2cm}{Spaced $k$-mers}
         & K-means & 0.126160 & 428.163587 & 3.500538 \\
         &  & Agglomerative & 0.122294 & 389.211763 & 3.418970 \\
         &  & K-Modes & -0.309347 & 4.658780 & 1.068371 \\
        \cmidrule{2-6} 
        & \multirow{3}{0.8cm}{PWM2Vec}
         & K-means & 0.061646 & 271.209939 & 3.586904 \\
         &  & Agglomerative & 0.019724 & 229.333382 & 4.163685 \\
         &  & K-Modes & -0.261891 & 1.060538 & 1.035597 \\
        \midrule
         \multirow{9}{1.2cm}{Isolation} & 
         \multirow{3}{0.8cm}{Spike2Vec}
        & K-means & 0.015248 & 58.296491 & 2.205289 \\
         &  & Agglomerative & 0.100906 & 66.592869 & 2.267826 \\
         &  & K-Modes & -0.406924 &  0.230669 & 1.709086 \\
         \cmidrule{2-6} 
         & \multirow{3}{1.2cm}{Spaced $k$-mers}
        & K-means & 0.894164 & 49.226179 & \textbf{0.064535} \\
         &  & Agglomerative & \textbf{0.894826} & 54.900365 & 2.231084 \\
         &  & K-Modes & -0.494872 &  3.841355 & 2.039936 \\
         \cmidrule{2-6} 
         & \multirow{3}{0.8cm}{PWM2Vec}
         & K-means & 0.427347 & 57.304634 & 2.029489 \\
         &  & Agglomerative & 0.892029 & 64.086967 & 1.472744 \\
         &  & K-Modes & -0.382076 &  0.146916 & 1.543540 \\
         \midrule
         
        \multirow{9}{1.2cm}{Modified Isolation (Ours)} & \multirow{3}{0.8cm}{Spike2Vec}
        & K-means & 0.100461 & \textbf{875.4463} & 1.249022 \\
         &  & Agglomerative & 0.258273 & 856.6747 & 1.088766 \\
         &  & K-Modes & -0.337161 & 0.404064 & 1.154124 \\
         \cmidrule{2-6} 
        & \multirow{3}{1.2cm}{Spaced $k$-mers}
         & K-means & 0.298419 & 766.048709 & 1.279214 \\
         &  & Agglomerative & 0.845733 & 807.127025 & 1.059536 \\
         &  & K-Modes & -0.491851 &  48.420757 & 2.083230 \\
         \cmidrule{2-6} 
        & \multirow{3}{0.8cm}{PWM2Vec}
        & K-means & 0.191296 & 626.270386 & 1.425956 \\
         &  & Agglomerative & 0.295157 & 594.130597 & 1.278188 \\
         &  & K-Modes & -0.358505 & 0.255160 & 1.504934 \\
         \bottomrule
         \end{tabular}
}
 \caption{Clustering results on \textbf{Protein Subcellular} dataset for different evaluation metrics. The best values are shown in bold.}
    \label{tbl_results_clustering_protein_subcellular}
\end{table}

\begin{table}[h!]
\centering
\resizebox{0.49\textwidth}{!}{
 \begin{tabular}{@{\extracolsep{6pt}}p{1.2cm}llp{1.6cm}p{1.4cm}p{1.4cm}}
    \toprule
        \multirow{1}{*}{Kernel} & \multirow{1}{*}{Embeddings} & \multirow{1}{*}{Algo.} & \multirow{1}{*}{Silhouette $\uparrow$} & \multirow{1}{*}{Calinski $\uparrow$} & \multirow{1}{*}{Davies $\downarrow$}\\
        \midrule \midrule
         
        \multirow{9}{1.2cm}{Gaussian} & 
        \multirow{3}{0.8cm}{Spike2Vec}
         & K-means & 0.725709 & 2360.1012 & 0.803562 \\
         &  & Agglomerative & 0.728701 & \textbf{2401.6734} & 0.897672 \\
         &  & K-Modes & -0.724895 & 84.479174 & 1.082510 \\
        \cmidrule{2-6} 
        & \multirow{3}{1.2cm}{Spaced $k$-mers}
         & K-means & 0.671649 & 1758.287930 & 0.582548 \\
         &  & Agglomerative & 0.697097 & 1750.801213 & 0.535291 \\
         &  & K-Modes & -0.526676 & 70.972746 & 1.769435 \\
        \cmidrule{2-6} 
        & \multirow{3}{0.8cm}{PWM2Vec}
         & K-means & 0.691450 & 1435.243293 & 0.693299 \\
         &  & Agglomerative & 0.660800 & 1401.089635 & 0.818952 \\
         &  & K-Modes & -0.520338 & 94.359037 & 2.462654 \\
        \midrule
         \multirow{9}{1.2cm}{Isolation} & 
         \multirow{3}{0.8cm}{Spike2Vec}
        & K-means & 0.068962 & 122.328499 & 0.912677 \\
         &  & Agglomerative & 0.597965 & 130.992233 & 2.127489 \\
         &  & K-Modes & -0.594450 & 0.137340 & 2.041841 \\
         \cmidrule{2-6} 
          & \multirow{3}{1.2cm}{Spaced $k$-mers}
        & K-means & 0.926955 & 85.446835 & 1.213278 \\
         &  & Agglomerative & 0.935912 & 90.096434 & 0.796774 \\
         &  & K-Modes & -0.624410 &  0.397808 & 2.425513 \\
         \cmidrule{2-6} 
         & \multirow{3}{0.8cm}{PWM2Vec}
         & K-means & \textbf{0.942670} & 90.830309 & \textbf{0.032672} \\
         &  & Agglomerative & 0.940357 & 91.914837 & 0.929066 \\
         &  & K-Modes & -0.671301 &  0.143632 & 2.922499 \\
         \midrule

        \multirow{9}{1.2cm}{Modified Isolation (Ours)} & 
        \multirow{3}{0.8cm}{Spike2Vec}
        & K-means & 0.062111 & 191.373940 & 2.674008 \\
         &  & Agglomerative & 0.069198 & 185.944651 & 2.621923 \\
         &  & K-Modes & -0.530450 & 1.363172 & 1.438185 \\
         \cmidrule{2-6} 
        & \multirow{3}{1.2cm}{Spaced $k$-mers}
         & K-means & 0.139867 & 689.139460 & 1.760740 \\
         &  & Agglomerative & 0.134961 & 688.940686 & 1.658960 \\
         &  & K-Modes & -0.461139 & 0.695404 & 1.294385 \\
         \cmidrule{2-6} 
        & \multirow{3}{0.8cm}{PWM2Vec}
        & K-means & 0.078648 & 322.821055 & 2.010462 \\
         &  & Agglomerative & 0.058910 & 312.395730 & 2.072378 \\
         &  & K-Modes & -0.377294 & 1.832262 & 1.199501 \\
         \bottomrule
         \end{tabular}
}
 \caption{Clustering results on \textbf{GISAID} dataset for different evaluation metrics. The best values are shown in bold.}
    \label{tbl_results_clustering_GISAID}
\end{table}

\begin{table}[h!]
\centering
\resizebox{0.49\textwidth}{!}{
 \begin{tabular}{@{\extracolsep{6pt}}p{1.2cm}llp{1.6cm}p{1.4cm}p{1.4cm}}
    \toprule
        \multirow{1}{*}{Kernel} & \multirow{1}{*}{Embeddings} & \multirow{1}{*}{Algo.} & \multirow{1}{*}{Silhouette $\uparrow$} & \multirow{1}{*}{Calinski $\uparrow$} & \multirow{1}{*}{Davies $\downarrow$}\\
        \midrule \midrule
         \multirow{9}{1.2cm}{Gaussian} & \multirow{3}{0.8cm}{Spike2Vec}
         & K-means & 0.845873 & 100.442214 &  0.099883 \\
         &  & Agglomerative & 0.849057 & 121.928330 &  0.097623 \\
         &  & K-Modes & -0.616155 & 0.680251 & 12.304512 \\
        \cmidrule{2-6} 
        & \multirow{3}{1.2cm}{Spaced $k$-mers}
         & K-means & 0.849962 & 110.207433 &  0.097384 \\
         &  & Agglomerative & 0.851768 & 122.335197 &  \textbf{0.096128} \\
         &  & K-Modes & -0.710487 & 0.595153 & 12.299829 \\
        \cmidrule{2-6}
        & \multirow{3}{0.8cm}{PWM2Vec}
         & K-means & 0.906833 & 1216.9008 & 0.950717 \\
         &  & Agglomerative & 0.907149 & 1242.8203 & 1.207521 \\
         &  & K-Modes & -0.783775 & 19.695747 & 1.991363 \\
        \midrule
         \multirow{9}{1.2cm}{Isolation} & \multirow{3}{0.8cm}{Spike2Vec}
        & K-means & 0.266999 & 85.054216 & 2.373702  \\
         &  & Agglomerative & 0.142199 & 95.157338 & 2.552974 \\
         &  & K-Modes & -0.398666 &  0.214001 & 1.951997 \\
         \cmidrule{2-6} 
         & \multirow{3}{1.2cm}{Spaced $k$-mers}
        & K-means & 0.059934 & 79.041871 & 2.562520 \\
         &  & Agglomerative & 0.045584 & 86.883693 & 2.917264 \\
         &  & K-Modes & -0.361089 &  0.390651 & 1.595999 \\
         \cmidrule{2-6} 
         & \multirow{3}{0.8cm}{PWM2Vec}
         & K-means & 0.493470 & 63.606950 & 1.599440 \\
         &  & Agglomerative & \textbf{0.928548} & 93.383507 & 0.510479 \\
         &  & K-Modes & -0.535966 &  6.971815 & 1.929305 \\
         \midrule
        \multirow{9}{1.2cm}{Modified Isolation (Ours)} & \multirow{3}{0.8cm}{Spike2Vec}
        & K-means & 0.326686 & 16050.3293 & 0.781717  \\
         &  & Agglomerative & 0.217707 & 15024.6706 & 1.035089 \\
         &  & K-Modes & -0.527625 & 6.309529 & 1.787152 \\
         \cmidrule{2-6} 
        & \multirow{3}{1.2cm}{Spaced $k$-mers}
         & K-means & 0.445578 & 4740.6237 & 0.496505 \\
         &  & Agglomerative & 0.414100 & 4529.5243 & 0.507057 \\
         &  & K-Modes & -0.618379 &  0.269577 & 2.518221 \\
         \cmidrule{2-6} 
        & \multirow{3}{0.8cm}{PWM2Vec}
        & K-means & 0.906386 & 85715.7221 & 0.201059 \\
         &  & Agglomerative & 0.906386 & \textbf{85715.7222} & 0.201059 \\
         &  & K-Modes & -0.486018 & 0.072697 & 2.195824 \\
         \bottomrule
         \end{tabular}
}
 \caption{Clustering results on \textbf{Nucleotide} dataset for different evaluation metrics. The best values are shown in bold.}
    \label{tbl_results_clustering_humna_dna}
\end{table}

\section{Statistical Significance of t-SNE Results}\label{sec_stats_tsne_appendix}
Since we use $1000$ iterations for tSNE computation, the reported results using neighborhood agreement and trustworthiness are statistically significant because of the convergence of t-SNE (since it is an iterative process, it converges every time within $1000$ iterations). Moreover, the neighborhood agreement and trustworthiness results for different neighbors $k$ (ranging from 1 to $100$) agreed with multiple runs of t-SNE (each with $1000$ iterations). Hence we report average results of $5$ t-SNE runs. The standard deviation values for both evaluation metrics were very small (i.e. $<0.002$). Based on the average and standard deviation results (of $5$ runs), we performed student $t$-test and the $p$-values were $<0.05$ in all cases.

\begin{table}[h!]
\centering
\resizebox{0.49\textwidth}{!}{
    \begin{tabular}{c@{\extracolsep{4pt}}ccc}
      \toprule
      
      \multirow{1}{*}{Dataset} & \multirow{1}{*}{Best Performing} & \multirow{1}{*}{Worst Performing} \\
      \midrule \midrule
      \multirow{1}{*}{Protein Subcellular} &  Random Walk & Random  \\
      \cmidrule{2-3}
      \multirow{1}{*}{GISAID} & Random Walk &  Random  \\
      \cmidrule{2-3}
      \multirow{1}{*}{Nucleotide} & Random Walk & Random \\
      \bottomrule
    \end{tabular}
    }
    \caption{Recommendation for initialization method based on the summary of performance on different datasets.}
    \label{tbl_recommended_result_summary}
\end{table}

\end{document}